\documentclass{article}

\usepackage[nonatbib,final]{neurips_2022}

\usepackage[utf8]{inputenc} %
\usepackage[T1]{fontenc}    %
\usepackage{hyperref}       %
\usepackage{url}            %
\usepackage{booktabs}       %
\usepackage{amsfonts}       %
\usepackage{nicefrac}       %
\usepackage{microtype}      %
\usepackage{xcolor}         %

\usepackage{graphicx,multirow}
\usepackage{amsmath}
\usepackage{amssymb}
\usepackage[export]{adjustbox}
\usepackage{siunitx} %
\usepackage{wrapfig}

\usepackage{placeins}
\usepackage{graphbox} %
\usepackage{breqn}
\usepackage{multirow}
\usepackage{mathtools}
\usepackage{bbm}
\usepackage{tabularx}
\usepackage{booktabs}
\usepackage{caption}
\usepackage{subcaption}
\usepackage[numbers]{natbib}

\usepackage{pifont}%
\newcommand{\cmark}{\ding{51}}%
\newcommand{\xmark}{\ding{55}}%

\newcommand{\sampling}{\xi_k}
\newcommand{\comment}[1]{}
\newcommand{\cf}{\emph{cf. }}
\newcommand{\eg}{\emph{e.g. }}

\newcommand{\diffusionmodel}{\emph{RDM}\,}
\newcommand{\diffusionmodels}{\emph{RDMs}\,}
\newcommand{\oidiffusion}{\emph{RDM-OI}\,}
\newcommand{\indiffusion}{\emph{RDM-IN}\,}
\newcommand{\wadiffusion}{\emph{RDM-WA}\,}
\newcommand{\cocodiffusion}{\emph{RDM-COCO}\,}
\newcommand{\oiindiffusion}{\emph{RDM-OI/IN}\,}
\newcommand{\inoidiffusion}{\emph{RDM-IN/OI}\,}
\newcommand{\armodel}{\emph{RARM}\,}

\title{Semi-Parametric Neural Image Synthesis}

\author{%
  Andreas Blattmann\thanks{The first two authors contributed equally to this work.} \quad Robin Rombach$^*$ \quad Kaan Oktay \quad Jonas M\"uller \quad Bj\"orn Ommer \\
  LMU Munich, MCML \& IWR, Heidelberg University, Germany\\
}

\providecommand{\imwidth}{}

\providecommand{\impath}[1]{}
\providecommand{\impaths}[1]{}
\providecommand{\impatha}[1]{}
\providecommand{\impathb}[1]{}
\providecommand{\impathc}[1]{}
\providecommand{\impathd}[1]{}
\providecommand{\impathe}[1]{}

\newcommand{\traindatabaseablation}{
\renewcommand{\impath}[1]{img/train_database_ablation/##1}
\begin{figure}[htbp]
\vspace{-1em}
\includegraphics[width=\textwidth]{\impath{metrics_over_trainsteps}}
\vspace{-1em}
\caption{\label{fig:dataset_ablation_metrics} Comparing performance metrics of \diffusionmodels with different train databases $\mathcal{D}_{\text{\tiny train}}$ with those of an \emph{LDM} baseline on the dogs-subset of ImageNet~\cite{imagenet}; we find that having a database of diverse visual instance from visual domains similar to the train dataset $\mathcal{X}$ (as \diffusionmodel\emph{-COCO}) improves performance upon fully-parametric baseline. Increasing the size of the database further boosts performance, leading to significant improvements of \diffusionmodels over the baseline despite having less trainable parameters.\vspace{-1em}}
\end{figure}
}

\newcommand{\retroguidingquantunwrapped}{
\renewcommand{\impath}[1]{img/retro_cfg_truncation/##1}
\begin{subfigure}[h]{0.45\textwidth}
\centering
\includegraphics[width=\imwidth]{\impath{prec_rec_adm_L}}
\includegraphics[width=\imwidth]{\impath{is_fid_adm_L}}
\caption{\label{fig:retro_guiding_wrapped} Assessing the effects of classifier free guidance.}
\end{subfigure}
}

\newcommand{\topmtruncationunwrapped}{
\renewcommand{\impath}[1]{img/topm_truncation/##1}
\begin{subfigure}[h]{0.45\textwidth}
\includegraphics[width=\imwidth]{\impath{prec-rec_L}}
\includegraphics[width=\imwidth]{\impath{is-fid_L}}
\caption{\label{fig:topm_trunc_wrapped} Quality-diversity trade-offs when applying top-m sampling.}
\end{subfigure}
}

\newcommand{\qualitydiversitymerged}{
\renewcommand{\imwidth}{0.48\textwidth}
\vspace{-1em}
\begin{figure*}[thbp]
\centering
\topmtruncationunwrapped
\retroguidingquantunwrapped
\caption{Analysis of the quality-diversity trade-offs when applying top-m sampling and classifier-free guidance. \vspace{-2.5em}}
\end{figure*}
}

\newcommand{\wrappedgeneralization}{

\renewcommand{\imwidth}{0.32\textwidth}
\renewcommand{\impath}[1]{img/generalization_quant/##1}
\begin{wrapfigure}{r}{.32\textwidth}
\vspace{-2.5em}
\includegraphics[width=\imwidth]{\impath{coco_in_new}}
\caption{\label{fig:generalization_quant} We observe that the number of neighbors $k_{\text{train}}$ retrieved during training significantly impacts the generalization abilities of \diffusionmodel. See Sec.~\ref{subsec:conditional}. \vspace{-2em}}
\end{wrapfigure}
}

\newcommand{\datasetcomplexity}{
\renewcommand{\imwidth}{0.325\textwidth}
\renewcommand{\impath}[1]{img/dset_complexity/##1}
\begin{figure}[t]
\vspace{-2.5em}
\includegraphics[width=0.975\textwidth]{\impath{all_combined_mono.jpg}}
\caption{\label{fig:dataset_complexity} Assessing our approach when increasing dataset complexity as in Sec.~\ref{subsec:dataset_complexity}.
We observe that performance-gaps between semi- and fully-parametric models increase for more complex datasets. %as indicated by the shaded areas. %Remarkably, the recall scores of of our proposed models \emph{increase} with increasing complexity, while strongly decreasing for the baseline models.
\vspace{-2em}}
\end{figure}

}

\newcommand{\modelfigure}{
\begin{figure}[t]
\centering
\includegraphics[width=\textwidth]{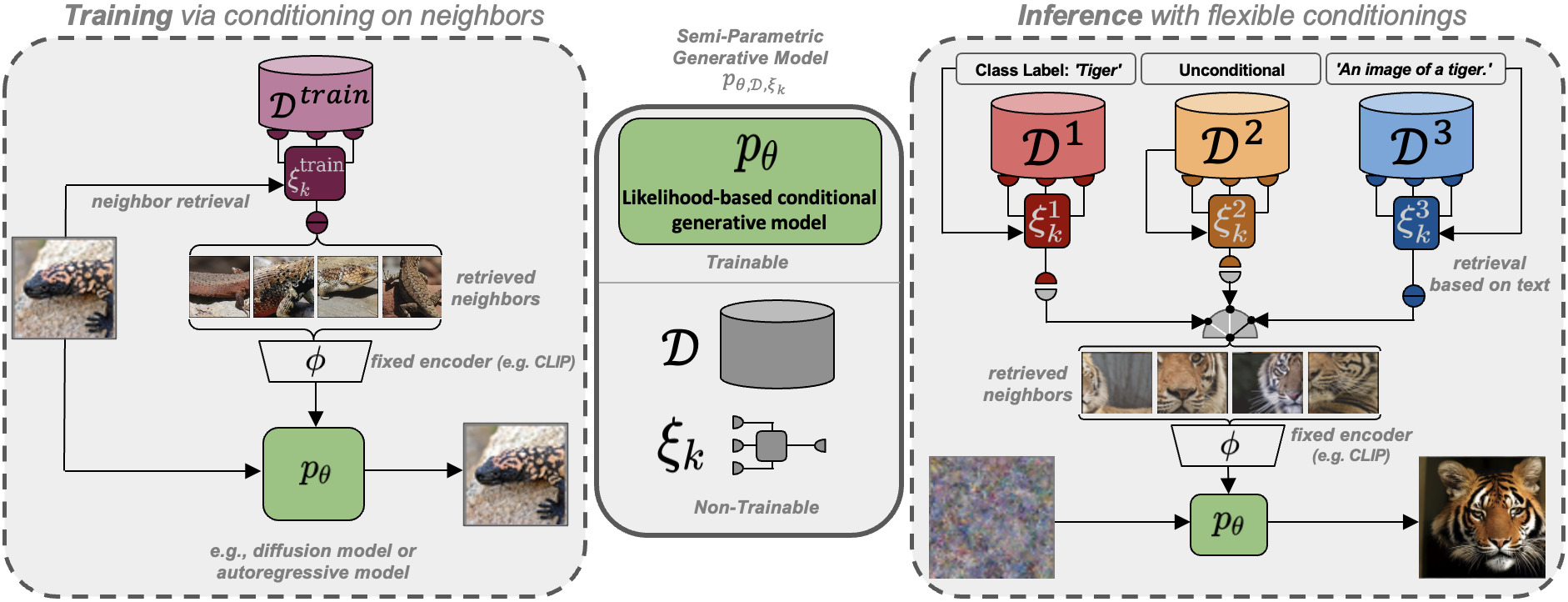}
\caption{\label{fig:model} A semi-parametric generative model consists of a trainable conditional generative model (decoding head) $p_\theta(x \vert \cdot)$, an external database $\mathcal{D}$ containing visual examples and a sampling strategy $\xi_k$ to obtain a subset $\mathcal{M}_{\mathcal{D}}^{(k)} \subseteq \mathcal{D}$, which serves as conditioning for $p_\theta$. During training, $\xi_k$ retrieves the nearest neighbors of each target example from $\mathcal{D}$, such that $p_\theta$ only needs to learn to compose consistent scenes based on $\mathcal{M}_{\mathcal{D}}^{(k)}$, \cf Sec~\ref{sec:model_training}. During inference, we can exchange $\mathcal{D}$ and $\xi_k$, thus resulting in flexible sampling capabilities % such as post-hoc conditioning an unconditional models on class labels or text prompts (right).}%
such as post-hoc conditioning on class labels ($\xi_k^1$) or text prompts ($\xi_k^3$), \cf Sec.~\ref{subsec:model_inference}, and zero-shot stylization, \cf Sec.~\ref{subsec:stylization}.\vspace{-2em}}
\end{figure}
}

\newcommand{\nnsexamples}{
\renewcommand{\imwidth}{.06\textwidth}
\renewcommand{\impath}[1]{img/clip_nns/##1}
\setlength{\tabcolsep}{2pt}
\renewcommand{\arraystretch}{0.0}
\begin{figure}[htbp]
\vspace{-0.5em}
\centering
\begin{tabular}{cc}
\toprule
 $x$ & $\xi_{15}(x)$ \\
\midrule
\includegraphics[width=\imwidth]{\impath{bald_eagle_ex}} &
\includegraphics[width=.915\textwidth]{\impath{bald_eagle_nns}} \\

\includegraphics[width=\imwidth]{\impath{german_shepperd_ex}} &
\includegraphics[width=.915\textwidth]{\impath{german_shepperd_nns}} \\
\bottomrule
\end{tabular}
\vspace{-.5em}
\caption{\label{fig:example_nns} $k=15$ nearest neighbors from $\mathcal{D}$ for a given query $x$ when parameterizing $d(x, \cdot)$ with CLIP~\cite{clip}. \vspace{-1.5em}
}
\end{figure}
}

\newcommand{\cliptextquerytwo}{
\begin{figure}[!t]
\renewcommand{\imwidth}{0.11\textwidth}
\renewcommand{\impath}[1]{img/text-guiding/##1}
\setlength{\tabcolsep}{0pt}
\centering
\begin{tabular}{c @{\hspace{3pt}} c@{\hspace{0pt}}c c@{\hspace{0pt}}c c@{\hspace{0pt}}c c@{\hspace{0pt}}c}
\toprule
 &\multicolumn{2}{c}{\parbox[c]{.22\textwidth}{\centering \vspace{-.8em} \tiny\emph{'A purple salamander in the grass.'}}} & \multicolumn{2}{c}{\parbox[c]{.22\textwidth}{\centering \vspace{-.8em}\tiny\emph{'A zebra-skinned panda.'}}} & \multicolumn{2}{c}{\shortstack{\tiny\emph{'A teddy bear} \\ \tiny\emph{ riding a motorcycle.'}}} & \multicolumn{2}{c}{\shortstack{\tiny\emph{'Image of a monkey} \\ \tiny\emph{ with the fur of a leopard.'}}}  \\
\midrule
\multirow{2}{*}{\shortstack{\scriptsize \emph{Text repr.} \\ \scriptsize \emph{only}}}&
\includegraphics[width=\imwidth]{\impath{purple_salamander_in_the_grass_1}} &
\includegraphics[width=\imwidth]{\impath{purple_salamander_in_the_grass_2}} &
\includegraphics[width=\imwidth]{\impath{zebra_panda_1}} &
\includegraphics[width=\imwidth]{\impath{zebra_panda_2}} &
\includegraphics[width=\imwidth]{\impath{teddy_bear_riding_motorcycle_1}} &
\includegraphics[width=\imwidth]{\impath{teddy_bear_riding_motorcycle_2}} &
\includegraphics[width=\imwidth]{\impath{monkey_leopard_4}} &
\includegraphics[width=\imwidth]{\impath{monkey_leopard_2}} \\[-3pt]

&
\includegraphics[width=\imwidth]{\impath{purple_salamander_in_the_grass_3}} &
\includegraphics[width=\imwidth]{\impath{purple_salamander_in_the_grass_4}} &
\includegraphics[width=\imwidth]{\impath{zebra_panda_3}} &
\includegraphics[width=\imwidth]{\impath{zebra_panda_4}} &
\includegraphics[width=\imwidth]{\impath{teddy_bear_riding_motorcycle_3}} &
\includegraphics[width=\imwidth]{\impath{teddy_bear_riding_motorcycle_4}} &
\includegraphics[width=\imwidth]{\impath{monkey_leopard_3}} &
\includegraphics[width=\imwidth]{\impath{monkey_leopard_5}} \\
\midrule

\parbox[c]{0.065\textwidth}{\shortstack{ \scriptsize \emph{Text repr. } \\ \scriptsize \emph{and NNs}}}&
\multicolumn{1}{m{\imwidth}}{\includegraphics[width=\imwidth]{\impath{no_rep_clip_salam1}}} &
\multicolumn{1}{m{\imwidth}}{\includegraphics[width=\imwidth]{\impath{no_rep_clip_salam2}}} &
\multicolumn{1}{m{\imwidth}}{\includegraphics[width=\imwidth]{\impath{no_rep_clip_panda1}}} &
\multicolumn{1}{m{\imwidth}}{\includegraphics[width=\imwidth]{\impath{no_rep_clip_panda2}}} &
\multicolumn{1}{m{\imwidth}}{\includegraphics[width=\imwidth]{\impath{no_rep_clip_teddy1}}} &
\multicolumn{1}{m{\imwidth}}{\includegraphics[width=\imwidth]{\impath{no_rep_clip_teddy2}}} &
\multicolumn{1}{m{\imwidth}}{\includegraphics[width=\imwidth]{\impath{no_rep_clip_monkey1}}} &
\multicolumn{1}{m{\imwidth}}{\includegraphics[width=\imwidth]{\impath{no_rep_clip_monkey2}}} \\

\midrule
\scriptsize \emph{NNs only } &
\multicolumn{1}{m{\imwidth}}{\includegraphics[width=\imwidth]{\impath{only_nns_salam1}}} &
\multicolumn{1}{m{\imwidth}}{\includegraphics[width=\imwidth]{\impath{only_nns_salam2}}} &
\multicolumn{1}{m{\imwidth}}{\includegraphics[width=\imwidth]{\impath{only_nns_panda1}}} &
\multicolumn{1}{m{\imwidth}}{\includegraphics[width=\imwidth]{\impath{only_nns_panda2}}} &
\multicolumn{1}{m{\imwidth}}{\includegraphics[width=\imwidth]{\impath{only_nns_teddy1}}} &
\multicolumn{1}{m{\imwidth}}{\includegraphics[width=\imwidth]{\impath{only_nns_teddy2}}} &
\multicolumn{1}{m{\imwidth}}{\includegraphics[width=\imwidth]{\impath{only_nns_monkey1}}} &
\multicolumn{1}{m{\imwidth}}{\includegraphics[width=\imwidth]{\impath{only_nns_monkey2}}} \\

\bottomrule

\end{tabular}

\caption{\label{fig:clipretro} As we retrieve nearest neighbors in the shared text-image space provided by CLIP, we can use text prompts as queries for exemplar-based synthesis. We observe our  \comment{ImageNet }\diffusionmodel to readily generalize to unseen and fictional text prompts when building the set of retrieved neighbors by directly conditioning on the CLIP text encoding $\phi_{\text{\tiny CLIP}}(c_{\text{\tiny text}})$ (top row). When using $\phi_{\text{\tiny CLIP}}(c_{\text{\tiny text}})$ together with its $k-1$ nearest neighbors from the retrieval database (middle row) or the $k$ nearest neighbors alone without the text representation, the model does not show these generalization capabilities (bottom row). \vspace{-2em}}
\end{figure}
}

\newcommand{\knnwrapped}{
\begin{wrapfigure}{r}{.45\textwidth}
\vspace{-1em}
\renewcommand{\imwidth}{0.22\textwidth}
\renewcommand{\impath}[1]{img/n_nns_train/##1}
\captionsetup[figure]{font=footnotesize,skip=0pt}
\includegraphics[width=\imwidth]{\impath{is_fid_in}}
\includegraphics[width=\imwidth]{\impath{prec_rec_in}}
\caption{\label{fig:knn_wrapped} Effect of $k_{\text{\scriptsize train}}$.\vspace{-1em}}
\end{wrapfigure}
}

\newcommand{\stylizer}{
\renewcommand{\imwidth}{.06\textwidth}
\renewcommand{\impatha}[1]{img/stylization/stag/##1}
\renewcommand{\impathb}[1]{img/stylization/fruit_basket/##1}
\renewcommand{\impathc}[1]{img/stylization/woman_piano/##1}
\renewcommand{\impathd}[1]{img/stylization/set_table/##1}
\setlength{\tabcolsep}{1pt}
\begin{wrapfigure}{r}{.55\textwidth}
\vspace{-1em}
\centering
\begin{tabular}{c@{\hspace{2pt}}c@{\hspace{0pt}}c c@{\hspace{0pt}}c c@{\hspace{0pt}}c c@{\hspace{0pt}}c}
\toprule
&\multicolumn{2}{c}{\parbox[c]{.1\textwidth}{\centering \vspace{-.8em} \scriptsize\emph{'A stag.'}}} &
\multicolumn{2}{c}{\shortstack{\scriptsize\emph{'A basket } \\
\scriptsize\emph{full of fruits.'}}}&
\multicolumn{2}{c}{\shortstack{\scriptsize\emph{'A woman } \\
\scriptsize\emph{playing piano.'}}} &
\multicolumn{2}{c}{\parbox[c]{.1\textwidth}{\centering \vspace{-.8em} \scriptsize\emph{'A table set.'}}} \\
\midrule
\multirow{2}{*}[0.03\textwidth]{\rotatebox[origin=c]{90}{$\mathcal{D}_{\text{\tiny style}}$}} &
{\includegraphics[width=\imwidth]{\impatha{style1}}} &
{\includegraphics[width=\imwidth]{\impatha{style2}}} &
{\includegraphics[width=\imwidth]{\impathb{style1}}} &
{\includegraphics[width=\imwidth]{\impathb{style3}}} &
{\includegraphics[width=\imwidth]{\impathc{style1}}} &
{\includegraphics[width=\imwidth]{\impathc{style2}}} &
{\includegraphics[width=\imwidth]{\impathd{style5}}} &
{\includegraphics[width=\imwidth]{\impathd{style1}}} \\ [-3pt]

&
{\includegraphics[width=\imwidth]{\impatha{style5}}} &
{\includegraphics[width=\imwidth]{\impatha{style3}}} &
{\includegraphics[width=\imwidth]{\impathb{style4}}} &
{\includegraphics[width=\imwidth]{\impathb{style5}}} &
{\includegraphics[width=\imwidth]{\impathc{style3}}} &
{\includegraphics[width=\imwidth]{\impathc{style5}}}&
{\includegraphics[width=\imwidth]{\impathd{style2}}} &
{\includegraphics[width=\imwidth]{\impathd{style4}}} \\

\midrule

\multirow{2}{*}[0.03\textwidth]{\rotatebox[origin=c]{90}{$\mathcal{D}_{\text{\tiny train}}$}} &
{\includegraphics[width=\imwidth]{\impatha{orig1}}} &
{\includegraphics[width=\imwidth]{\impatha{orig2}}} &
{\includegraphics[width=\imwidth]{\impathb{orig1}}}&
{\includegraphics[width=\imwidth]{\impathb{orig2}}} &
{\includegraphics[width=\imwidth]{\impathc{orig1}}} &
{\includegraphics[width=\imwidth]{\impathc{orig2}}} &
{\includegraphics[width=\imwidth]{\impathd{orig1}}} &
{\includegraphics[width=\imwidth]{\impathd{orig2}}} \\[-3pt]

&
{\includegraphics[width=\imwidth]{\impatha{orig3}}} &
{\includegraphics[width=\imwidth]{\impatha{orig4}}} &
{\includegraphics[width=\imwidth]{\impathb{orig3}}} &
{\includegraphics[width=\imwidth]{\impathb{orig5}}} &
{\includegraphics[width=\imwidth]{\impathc{orig3}}} &
{\includegraphics[width=\imwidth]{\impathc{orig4}}} &
{\includegraphics[width=\imwidth]{\impathd{orig3}}} &
{\includegraphics[width=\imwidth]{\impathd{orig4}}} \\

\bottomrule
\end{tabular}
\caption{\label{fig:stylize} Zero-shot text-guided stylization with our ImageNet-\diffusionmodel. Best viewed when zoomed in. \vspace{-1.5em}}
\end{wrapfigure}
}

\newcommand{\generator}{
\renewcommand{\imwidth}{.06\textwidth}
\renewcommand{\impatha}[1]{img/flow_exp/noflow/bear/##1}
\renewcommand{\impathb}[1]{img/flow_exp/noflow/sala/##1}
\renewcommand{\impathc}[1]{img/flow_exp/flow/bear/##1}
\renewcommand{\impathd}[1]{img/flow_exp/flow/sala/##1}
\setlength{\tabcolsep}{1pt}
\begin{wrapfigure}{r}{.3\textwidth}
\vspace{-1em}
\centering
\begin{tabular}{c@{\hspace{2pt}}c@{\hspace{0pt}}c c@{\hspace{0pt}}c}
&\multicolumn{2}{c}{\parbox[c]{.1\textwidth}{\centering \vspace{-.8em} \scriptsize\emph{'A brown bear.'}}} &
\multicolumn{2}{c}{\parbox[c]{.1\textwidth}{\centering \vspace{-.8em} \scriptsize\emph{'A yellow bird.'}}} \\

\midrule
\multirow{2}{*}{\rotatebox[origin=c]{90}{\tiny{no prior}}} &
\multicolumn{1}{m{\imwidth}}{\includegraphics[width=\imwidth]{\impatha{0000}}} &
\multicolumn{1}{m{\imwidth}}{\includegraphics[width=\imwidth]{\impatha{0001}}} &
\multicolumn{1}{m{\imwidth}}{\includegraphics[width=\imwidth]{\impathb{0022}}} &
\multicolumn{1}{m{\imwidth}}{\includegraphics[width=\imwidth]{\impathb{0023}}} \\ [-3pt]

&
\multicolumn{1}{m{\imwidth}}{\includegraphics[width=\imwidth]{\impatha{0002}}} &
\multicolumn{1}{m{\imwidth}}{\includegraphics[width=\imwidth]{\impatha{0003}}} &
\multicolumn{1}{m{\imwidth}}{\includegraphics[width=\imwidth]{\impathb{0024}}} &
\multicolumn{1}{m{\imwidth}}{\includegraphics[width=\imwidth]{\impathb{0032}}} \\

\midrule

\multirow{2}{*}{\rotatebox[origin=c]{90}{\tiny{flow prior}}} &
\multicolumn{1}{m{\imwidth}}{\includegraphics[width=\imwidth]{\impathc{0002}}} &
\multicolumn{1}{m{\imwidth}}{\includegraphics[width=\imwidth]{\impathc{0004}}} &
\multicolumn{1}{m{\imwidth}}{\includegraphics[width=\imwidth]{\impathd{0019}}} &
\multicolumn{1}{m{\imwidth}}{\includegraphics[width=\imwidth]{\impathd{0022}}} \\[-3pt]

&

\multicolumn{1}{m{\imwidth}}{\includegraphics[width=\imwidth]{\impathc{0008}}} &
\multicolumn{1}{m{\imwidth}}{\includegraphics[width=\imwidth]{\impathc{0011}}} &
\multicolumn{1}{m{\imwidth}}{\includegraphics[width=\imwidth]{\impathd{0028}}} &
\multicolumn{1}{m{\imwidth}}{\includegraphics[width=\imwidth]{\impathd{0030}}} \\

\bottomrule
\end{tabular}
\caption{\label{fig:generator} Text-to-image generalization needs a generative prior or retrieval. See Sec.~\ref{subsec:conditional}. \vspace{-1.0em}
}
\end{wrapfigure}
}

\newcommand{\ccsamples}{
\renewcommand{\imwidth}{.0575\textwidth}
\renewcommand{\impath}[1]{img/class_conditional/##1}
\setlength{\tabcolsep}{1pt}
\begin{figure}[htbp]
\centering
\vspace{-1em}
\begin{tabular}{c@{\hspace{0pt}}cc@{\hspace{0pt}}cc@{\hspace{0pt}}cc@{\hspace{0pt}}cc@{\hspace{0pt}}cc@{\hspace{0pt}}cc@{\hspace{0pt}}cc@{\hspace{0pt}}c}
\toprule
\multicolumn{2}{c}{\scriptsize\emph{'Tench'}} & \multicolumn{2}{c}{\scriptsize\emph{'Vulture'}} & \multicolumn{2}{c}{\scriptsize\emph{'Grey Fox'}} & \multicolumn{2}{c}{\scriptsize\emph{'Tiger'}} & \multicolumn{2}{c}{\scriptsize\emph{'Teddy Bear'}} & \multicolumn{2}{c}{\scriptsize\emph{'Moped'}} & \multicolumn{2}{c}{\scriptsize\emph{'Harvester'}} & \multicolumn{2}{c}{\scriptsize\emph{'Espresso'}} \\
\midrule
\includegraphics[width=\imwidth]{\impath{tench1}} &
\includegraphics[width=\imwidth]{\impath{tench7}} &
\includegraphics[width=\imwidth]{\impath{vulture1}} &
\includegraphics[width=\imwidth]{\impath{vulture2}} &
\includegraphics[width=\imwidth]{\impath{grey_fox1}} &
\includegraphics[width=\imwidth]{\impath{grey_fox3}} &
\includegraphics[width=\imwidth]{\impath{tiger1}} &
\includegraphics[width=\imwidth]{\impath{tiger6}} &
\includegraphics[width=\imwidth]{\impath{teddy1}} &
\includegraphics[width=\imwidth]{\impath{teddy5}} &
\includegraphics[width=\imwidth]{\impath{moped1}} &
\includegraphics[width=\imwidth]{\impath{moped3}} &
\includegraphics[width=\imwidth]{\impath{harvester1}} &
\includegraphics[width=\imwidth]{\impath{harvester4}} &
\includegraphics[width=\imwidth]{\impath{espresso1}} &
\includegraphics[width=\imwidth]{\impath{espresso4}} \\[-3pt]

\includegraphics[width=\imwidth]{\impath{tench2}} &
\includegraphics[width=\imwidth]{\impath{tench9}} &
\includegraphics[width=\imwidth]{\impath{vulture4}} &
\includegraphics[width=\imwidth]{\impath{vulture5}} &
\includegraphics[width=\imwidth]{\impath{grey_fox2}} &
\includegraphics[width=\imwidth]{\impath{grey_fox6}} &
\includegraphics[width=\imwidth]{\impath{tiger2}} &
\includegraphics[width=\imwidth]{\impath{tiger5}} &
\includegraphics[width=\imwidth]{\impath{teddy4}} &
\includegraphics[width=\imwidth]{\impath{teddy2}} &
\includegraphics[width=\imwidth]{\impath{moped2}} &
\includegraphics[width=\imwidth]{\impath{moped4}} &
\includegraphics[width=\imwidth]{\impath{harvester3}} &
\includegraphics[width=\imwidth]{\impath{harvester5}} &
\includegraphics[width=\imwidth]{\impath{espresso3}} &
\includegraphics[width=\imwidth]{\impath{espresso5}} \\[-3pt]

\bottomrule
\end{tabular}
\caption{\label{fig:zero_shot_cc}
\diffusionmodel can be used for class-conditional generation on ImageNet despite being trained without class labels. %, see Sec.~\ref{}
To achieve this during inference, we compute a pool of nearby visual instances from the database $\mathcal{D}$ for each class label based on its textual description,
and combine it with its $k-1$ nearest neighbors as conditioning. \vspace{-2.2em}
}
\end{figure}
}

\newcommand{\sampleswithnns}{
\begin{figure}[t]
\renewcommand{\imwidth}{.0639\textwidth}
\renewcommand{\impath}[1]{img/samples_nns_d_nns_q/##1}
\renewcommand{\impatha}[1]{img/samples_nns_d_nns_q/ar/##1}
\setlength{\tabcolsep}{1pt}
\begin{tabularx}{\textwidth}{c@{\hspace{4pt}}ccccccccc | cc | cc}
\toprule
& \multicolumn{11}{c}{\diffusionmodel} & \multicolumn{2}{|c}{\armodel} \\[3pt]
& \multicolumn{9}{c}{\emph{\scriptsize{ImageNet~\cite{imagenet}}}} & \multicolumn{2}{|c}{\emph{\scriptsize{FFHQ~\cite{stylegan1}}}} & \multicolumn{2}{|c}{\emph{\scriptsize{ImageNet}}} \\
\midrule
$\mathcal{M}_{\mathcal{D}}^{(k)}(\tilde{x})$ &
\multicolumn{1}{m{\imwidth}}{\shortstack{\includegraphics[width=\imwidth]{\impath{d_nns_img0-1}} \\[-2.8pt]
                                         \includegraphics[width=\imwidth]{\impath{d_nns_img0-2}}}}&
\multicolumn{1}{m{\imwidth}}{\shortstack{\includegraphics[width=\imwidth]{\impath{d_nns_img3-1}} \\[-2.8pt]
                                         \includegraphics[width=\imwidth]{\impath{d_nns_img3-2}}}}&
\multicolumn{1}{m{\imwidth}}{\shortstack{\includegraphics[width=\imwidth]{\impath{d_nns_img4-1}} \\[-2.8pt]
                                         \includegraphics[width=\imwidth]{\impath{d_nns_img4-2}}}}&
\multicolumn{1}{m{\imwidth}}{\shortstack{\includegraphics[width=\imwidth]{\impath{d_nns_img6-1}} \\[-2.8pt]
                                         \includegraphics[width=\imwidth]{\impath{d_nns_img6-2}}}}&
\multicolumn{1}{m{\imwidth}}{\shortstack{\includegraphics[width=\imwidth]{\impath{d_nns_img9-1}} \\[-2.8pt]
                                         \includegraphics[width=\imwidth]{\impath{d_nns_img9-2}}}}&
\multicolumn{1}{m{\imwidth}}{\shortstack{\includegraphics[width=\imwidth]{\impath{d_nns_img11-1}} \\[-2.8pt]
                                         \includegraphics[width=\imwidth]{\impath{d_nns_img11-2}}}}&
\multicolumn{1}{m{\imwidth}}{\shortstack{\includegraphics[width=\imwidth]{\impath{d_nns_img12-1}} \\[-2.8pt]
                                         \includegraphics[width=\imwidth]{\impath{d_nns_img12-2}}}}&
\multicolumn{1}{m{\imwidth}}{\shortstack{\includegraphics[width=\imwidth]{\impath{d_nns_img13-1}} \\[-2.8pt]
                                         \includegraphics[width=\imwidth]{\impath{d_nns_img13-2}}}}&
\multicolumn{1}{m{\imwidth}}{\shortstack{\includegraphics[width=\imwidth]{\impath{d_nns_img14-1}} \\[-2.8pt]
                                         \includegraphics[width=\imwidth]{\impath{d_nns_img14-2}}}}  &
\multicolumn{1}{ | m{\imwidth}}{\shortstack{\includegraphics[width=\imwidth]{\impath{ffhq_nns_database_1-1}} \\[-2.8pt]
                                         \includegraphics[width=\imwidth]{\impath{ffhq_nns_database_1-2}}}}&
\multicolumn{1}{m{\imwidth}}{\shortstack{\includegraphics[width=\imwidth]{\impath{ffhq_nns_database2-1}} \\[-2.8pt]
                                         \includegraphics[width=\imwidth]{\impath{ffhq_nns_database2-2}}}} &
\multicolumn{1}{ | m{\imwidth}}{\shortstack{\includegraphics[width=\imwidth]{\impatha{dogs_nns_db1}} \\[-2.8pt]
                                         \includegraphics[width=\imwidth]{\impatha{dogs_nns_db2}}}}&
\multicolumn{1}{m{\imwidth}}{\shortstack{\includegraphics[width=\imwidth]{\impatha{water_nns_db1}} \\[-2.8pt]
                                         \includegraphics[width=\imwidth]{\impatha{water_nns_db2}}}}\\[-1pt]

\midrule
 &
\multicolumn{1}{m{\imwidth}}{\includegraphics[width=\imwidth]{\impath{img0}}}&
\multicolumn{1}{m{\imwidth}}{\includegraphics[width=\imwidth]{\impath{img3}}}&
\multicolumn{1}{m{\imwidth}}{\includegraphics[width=\imwidth]{\impath{img4}}}&
\multicolumn{1}{m{\imwidth}}{\includegraphics[width=\imwidth]{\impath{img6}}}&
\multicolumn{1}{m{\imwidth}}{\includegraphics[width=\imwidth]{\impath{img9}}}&
\multicolumn{1}{m{\imwidth}}{\includegraphics[width=\imwidth]{\impath{img11}}}&
\multicolumn{1}{m{\imwidth}}{\includegraphics[width=\imwidth]{\impath{img12}}}&
\multicolumn{1}{m{\imwidth}}{\includegraphics[width=\imwidth]{\impath{img13}}}&
\multicolumn{1}{m{\imwidth}}{\includegraphics[width=\imwidth]{\impath{img14}}}&
\multicolumn{1}{|m{\imwidth}}{\includegraphics[width=\imwidth]{\impath{ffhq_sample_1-1}}}&
\multicolumn{1}{m{\imwidth}}{\includegraphics[width=\imwidth]{\impath{ffhq_sample_2-3}}}&
\multicolumn{1}{|m{\imwidth}}{\includegraphics[width=\imwidth]{\impatha{dogs1}}}&
\multicolumn{1}{m{\imwidth}}{\includegraphics[width=\imwidth]{\impatha{water1}}} \\[-2.8pt]
\emph{Samples} &
\multicolumn{1}{m{\imwidth}}{\includegraphics[width=\imwidth]{\impath{img01}}}&
\multicolumn{1}{m{\imwidth}}{\includegraphics[width=\imwidth]{\impath{img31}}}&
\multicolumn{1}{m{\imwidth}}{\includegraphics[width=\imwidth]{\impath{img41}}}&
\multicolumn{1}{m{\imwidth}}{\includegraphics[width=\imwidth]{\impath{img61}}}&
\multicolumn{1}{m{\imwidth}}{\includegraphics[width=\imwidth]{\impath{img91}}}&
\multicolumn{1}{m{\imwidth}}{\includegraphics[width=\imwidth]{\impath{img111}}}&
\multicolumn{1}{m{\imwidth}}{\includegraphics[width=\imwidth]{\impath{img121}}}&
\multicolumn{1}{m{\imwidth}}{\includegraphics[width=\imwidth]{\impath{img131}}}&
\multicolumn{1}{m{\imwidth}}{\includegraphics[width=\imwidth]{\impath{img141}}}&
\multicolumn{1}{|m{\imwidth}}{\includegraphics[width=\imwidth]{\impath{ffhq_sample_1-2}}}&
\multicolumn{1}{m{\imwidth}}{\includegraphics[width=\imwidth]{\impath{ffhq_sample_2-2}}}&
\multicolumn{1}{|m{\imwidth}}{\includegraphics[width=\imwidth]{\impatha{dogs2}}}&
\multicolumn{1}{m{\imwidth}}{\includegraphics[width=\imwidth]{\impatha{water2}}} \\[-2.8pt]
 &
\multicolumn{1}{m{\imwidth}}{\includegraphics[width=\imwidth]{\impath{img02}}}&
\multicolumn{1}{m{\imwidth}}{\includegraphics[width=\imwidth]{\impath{img32}}}&
\multicolumn{1}{m{\imwidth}}{\includegraphics[width=\imwidth]{\impath{img42}}}&
\multicolumn{1}{m{\imwidth}}{\includegraphics[width=\imwidth]{\impath{img62}}}&
\multicolumn{1}{m{\imwidth}}{\includegraphics[width=\imwidth]{\impath{img92}}}&
\multicolumn{1}{m{\imwidth}}{\includegraphics[width=\imwidth]{\impath{img112}}}&
\multicolumn{1}{m{\imwidth}}{\includegraphics[width=\imwidth]{\impath{img122}}}&
\multicolumn{1}{m{\imwidth}}{\includegraphics[width=\imwidth]{\impath{img132}}}&
\multicolumn{1}{m{\imwidth}}{\includegraphics[width=\imwidth]{\impath{img142}}}&
\multicolumn{1}{|m{\imwidth}}{\includegraphics[width=\imwidth]{\impath{ffhq_sample_1-3}}}&
\multicolumn{1}{m{\imwidth}}{\includegraphics[width=\imwidth]{\impath{ffhq_sample_2-1}}}&
\multicolumn{1}{|m{\imwidth}}{\includegraphics[width=\imwidth]{\impatha{dogs3}}}&
\multicolumn{1}{m{\imwidth}}{\includegraphics[width=\imwidth]{\impatha{water3}}} \\[-1pt]

\midrule
\shortstack{\emph{NNs in } \\
            \emph{train set}
            \vspace{-1em}} &
\multicolumn{1}{m{\imwidth}}{\includegraphics[width=\imwidth]{\impath{img0_nns_less}}}&
\multicolumn{1}{m{\imwidth}}{\includegraphics[width=\imwidth]{\impath{img3_nns_less}}}&
\multicolumn{1}{m{\imwidth}}{\includegraphics[width=\imwidth]{\impath{img4_nns_less}}}&
\multicolumn{1}{m{\imwidth}}{\includegraphics[width=\imwidth]{\impath{img6_nns_less}}}&
\multicolumn{1}{m{\imwidth}}{\includegraphics[width=\imwidth]{\impath{img9_nns_less}}}&
\multicolumn{1}{m{\imwidth}}{\includegraphics[width=\imwidth]{\impath{img11_nns_less}}}&
\multicolumn{1}{m{\imwidth}}{\includegraphics[width=\imwidth]{\impath{img12_nns_less}}}&
\multicolumn{1}{m{\imwidth}}{\includegraphics[width=\imwidth]{\impath{img13_nns_less}}}&
\multicolumn{1}{m{\imwidth}}{\includegraphics[width=\imwidth]{\impath{img14_nns_less}}}&
\multicolumn{1}{|m{\imwidth}}{\includegraphics[width=\imwidth]{\impath{ffhq_nns_1}}}&
\multicolumn{1}{m{\imwidth}}{\includegraphics[width=\imwidth]{\impath{ffhq_nns_2}}}&
\multicolumn{1}{|m{\imwidth}}{\includegraphics[width=\imwidth]{\impatha{dogs1_nns_less}}}&
\multicolumn{1}{m{\imwidth}}{\includegraphics[width=\imwidth]{\impatha{water1_nns_less}}} \\

\bottomrule
\end{tabularx}
\caption{\label{fig:samples_with_nns} Samples from our unconditional models together with the sets of $\mathcal{M}_{\mathcal{D}}^{(k)}(\tilde{x})$ of retrieved neighbors for the pseudo query $\tilde{x}$, \cf Sec.~\ref{subsec:model_inference}, and nearest neighbors from the train set, measured in CLIP~\cite{clip} feature space. For ImageNet samples are generated with $m=0.01$, guidance with $s = 2.0$ and 100 DDIM steps for \diffusionmodel and $m=0.05$, guidance scale $s=3.0$ and top-$k=2048$ for \armodel. On FFHQ we use $s=1.0\,, m=0.1$.\vspace{-2em}}
\end{figure}}

\newcommand{\fpp}{
\begin{wrapfigure}{r}{.34\textwidth}
\vspace{-1em}
\renewcommand{\imwidth}{0.34\textwidth}
\renewcommand{\impath}[1]{img/##1}
\vspace{-.5em}
\includegraphics[width=\imwidth]{\impath{first_page_float_brokenx}}
\caption{\label{fig:fpp} Our semi-parametric model outperforms the unconditional SOTA model ADM~\cite{adm} on ImageNet~\cite{imagenet} and even reaches the class-conditional ADM (ADM w/ classifier), while reducing parameter count. $\vert\mathcal{D}\vert\colon$ Number of instances in database at inference; $\vert\theta\vert\colon$ Number of trainable parameters.\vspace{-1em}}
\vspace{-1em}
\end{wrapfigure}
}

\newcommand{\ingeneralization}{
\begin{tabular}{lcccccc}
\toprule
\textbf{Method} & \multicolumn{2}{c}{FID$\downarrow$} &\multicolumn{2}{c}{CLIP-FID$\downarrow$} & Precision$\uparrow$ & Recall$\uparrow$ \\
& \textit{\scriptsize{train}} & \textit{\scriptsize{val}} & \textit{\scriptsize{train}} & \textit{\scriptsize{val}} & & \\
\midrule
\indiffusion & 5.91 & 5.32 & 3.92 & 4.44 & 0.74 & 0.51 \\[3pt]
\oidiffusion & 12.28 & 11.31 & 4.09 & 4.59 & 0.69 & 0.55 \\[3pt]
\inoidiffusion & 17.23 & 16.82 & 8.86 & 9.75 & 0.52 & 0.60 \\[3pt]
\oiindiffusion & 10.81 & 12.01 & 3.84 & 4.41 & 0.81 & 0.39 \\[3pt]
\bottomrule
\end{tabular}
}

\newcommand{\cocotexttoimage}{

\begin{tabular}{lcccc}
\toprule
\textbf{Method} & FID$\downarrow$ &CLIP-FID$\downarrow$ & CLIP-score$\uparrow$& IS$\uparrow$\\
\midrule
LAFITE~\cite{zhou2021lafite} & 26.94 & - & - & \textbf{26.02} \\[3pt]
\indiffusion & 27.28 & 18.12 & 0.29 & 24.17 \\[3pt]
\oidiffusion & \textbf{22.08} & \textbf{13.16} & \textbf{0.30} & 24.31  \\[3pt]
\bottomrule
\end{tabular}%
}

\newcommand{\databasegeneralization}{
\begin{table}[htbp]
\centering
\vspace{-1em}
\resizebox{.47\textwidth}{!}{
\ingeneralization
}
\hfill
\resizebox{.52\textwidth}{!}{
\cocotexttoimage
}
\caption{\label{tab:database_generalization} \emph{Generalization to new databases}. Left: We train \diffusionmodels on ImageNet with OpenImages (\oidiffusion\!) and the train dataset itself (\indiffusion\!). By exchanging the train and inference databases between the two models we see that \oidiffusion which is trained with a database disjoint from the train set generalizes better to new inference databases. Right: Quantitative comparison against LAFITE~\cite{zhou2021lafite} on zero-shot text-to-image synthesis. \vspace{-2em}}
\end{table}
}

\newcommand{\uinmetrics}{
\begin{table}[htbp]
\vspace{-2em}
\centering
\begin{footnotesize}
\begin{adjustbox}{max width=\linewidth}
\footnotesize
\begin{tabular}{lccccccccl}
\toprule
\textbf{Method} & \multicolumn{2}{c}{FID$\downarrow$} & IS$\uparrow$ & \multicolumn{2}{c}{Precision$\uparrow$} & \multicolumn{2}{c}{Recall$\uparrow$} & $N_{\text{params}}$ & \\
 & \textit{\scriptsize{train}} & \textit{\scriptsize{val}} &  & \textit{\scriptsize{train}} & \textit{\scriptsize{val}} & \textit{\scriptsize{train}} & \textit{\scriptsize{val}} & &\\
\midrule
IC-GAN~\cite{ic-gan} & 18.17 & 15.60$^{*}$ & 59.00$^{*}$ & \underline{0.77} & \textbf{0.73} & 0.21 & 0.23  & 191M & conditioned on train set, add. aug.\\
ADM~\cite{adm} & 26.21 & 32.50$^{*}$ & 39.70 & 0.61& - & \underline{0.63} & -  & 554M & 250 steps\\
ADM-G~\cite{adm} & 33.03 & - & 32.92 & 0.56&- & \textbf{0.65} & - & 618M & 250 steps, c.g., s=1.0 \\
ADM-G~\cite{adm} & \underline{12.00} & - & \underline{95.41} & 0.76&- &  0.44 & - & 618M & 250 steps, c.g., s=10.0 \\
\midrule
\oidiffusion (ours) & 24.50 & 21.28  & 45.29 & 0.60& 0.54 & \textbf{0.65} & \textbf{0.66} & 400M & 100 steps, $m=0.1$\\
\oidiffusion (ours) & 19.08 & 16.89 & 62.78  & 0.57& 0.62 & 0.56 & \underline{0.57} & 400M & 100 steps, $m=0.01$\\
\oidiffusion (ours) & 13.22 & 12.29 & 70.64 & 0.72& 0.65 &  0.56 &  0.51 & 400M & 100 steps, c.f.g., $s=1.75$, $m=0.1$\\
\oidiffusion (ours) & 13.60 & 13.11 & 87.58   & \textbf{0.79}& \textbf{0.73} & 0.51 & 0.50  & 400M & 100 steps, c.f.g., $s=1.5$, $m=0.02$\\
\oidiffusion (ours) & 12.21 & \underline{11.31}& 77.93  & 0.75& \underline{0.69} &  0.55  & 0.55 & 400M & 100 steps, c.f.g., $s=1.5$, $m=0.05$\\
\indiffusion (ours) & \textbf{5.91} & \textbf{5.32} & \textbf{158.76} & 0.74 & 0.74  & 0.51 & 0.53 & 400M & 100 steps, c.f.g., $s=1.5$, $m=0.05$ \\
\midrule
\end{tabular}
\end{adjustbox}
\end{footnotesize}
\caption{\label{tab:uin_metrics} Comparison of \diffusionmodel with recent state-of-the-art methods for unconditional image generation on ImageNet~\cite{imagenet}. While \emph{c.f.g.} denotes classifier-free guidance with a scale parameter $s$ as proposed in~\cite{ho2021classifier}, \emph{c.g.} refers to classifier guidance~\cite{adm}, what requires a classifier pretrained on the noisy representations of diffusion models to be available. $^{*}$: numbers taken from~\cite{ic-gan}.\vspace{-3em}}
\end{table}
}

\newcommand{\wikiartmetrics}{
\begin{wraptable}{r}{.5\textwidth}
\centering
\setlength{\tabcolsep}{2pt}
\vspace{-1em}
\begin{adjustbox}{max width=.6\textwidth}
\begin{tabular}{lcccccc}
\toprule
\textbf{Method} & \multicolumn{2}{c}{FID$\downarrow$} &\multicolumn{2}{c}{Precision$\uparrow$} & \multicolumn{2}{c}{Recall$\uparrow$} \\[3pt]
\scriptsize{Backbone} & \textit{\scriptsize{I-V3}} & \textit{\scriptsize{CLIP}} & \textit{\scriptsize{I-V3}} & \textit{\scriptsize{CLIP}} & \textit{\scriptsize{I-V3}} & \textit{\scriptsize{CLIP}} \\[3pt]
\midrule
IC-GAN & 24.75 & 35.17 & 0.47 & 0.38 & 0.28  & 0.02\\[3pt]
\oidiffusion & \textbf{21.50} & \textbf{13.01} & \textbf{0.63} & \textbf{0.46} & \textbf{0.34} & \textbf{0.11} \\[3pt]
\bottomrule
\end{tabular}
\end{adjustbox}
\caption{\label{tab:wa_metrics} Performance metrics evaluated against examples from WikiArt for IC-GAN and \oidiffusion trained on ImageNet. During inference both models are conditioned on samples from the WikiArt database. \vspace{-1em}}
\end{wraptable}
}

\newcommand{\ffhqmetrics}{
\begin{wraptable}{r}{.4\textwidth}
\centering
\vspace{-.5em}
\begin{adjustbox}{max width=.4\textwidth}
\begin{tabular}{lccc}
\toprule
\textbf{Method} & CLIP-FID &  CLIP-Prec & CLIP-Rec \\[3pt]
\midrule
P-GAN~\cite{DBLP:journals/corr/abs-2111-01007} & 4.87 & - & - \\[3pt]
Style-GAN2~\cite{stylegan2} & 2.90 & - & - \\[3pt]
LDM~\cite{ldm} & \underline{2.12} & 0.81 & \textbf{0.48} \\[3pt]
LDM (equal $N_{\text{\tiny params}}$) & 2.63 & \underline{0.87}  & \underline{0.44} \\[3pt]
\oidiffusion & \textbf{1.92}& \textbf{0.93} & 0.35 \\[3pt]
\bottomrule
\end{tabular}
\end{adjustbox}
\caption{\label{tab:ffhq_metrics} Quantiative results on FFHQ~\cite{stylegan1}. \oidiffusion samples generated with $m=0.1$ and without classifier-free guidance.\vspace{-1.2em}}
\end{wraptable}
}

\newcommand{\hyperparams}{
\begin{table}[thbp]
\centering
\setlength{\tabcolsep}{5pt}

\begin{adjustbox}{max width=\linewidth}
\footnotesize
\begin{tabular}{lcccc}
\toprule
& \diffusionmodel$^{*}$ & \diffusionmodel$^{\dagger}$ & \diffusionmodel$^{\ddagger}$ & baseline \emph{LDM}$^{\ddagger}$  \\[3pt]
\midrule
Dataset & ImageNet (IN) & ImageNet & IN-subsets, \cf Tab.~\ref{tab:in_subsets} & IN-subsets, \cf Tab.~\ref{tab:in_subsets} \\[3pt]
$z$-shape & $64 \times 64 \times 3$ & $16 \times 16 \times 16$ & $64 \times 64 \times 3$ & $64 \times 64 \times 3$  \\[3pt]
$\vert \mathcal{Z} \vert$ & 8192 & KL  & 8192 & 8192  \\[3pt]
Diffusion steps &1000 & 1000 & 1000 & 1000 \\[3pt]
Noise Schedule & linear&linear&linear&linear\\[3pt]
Model Size &400M & 200M & 400M & 576M\\[3pt]
Channels & 192 & 192 & 192 & 224 \\[3pt]
Depth &2& 2 & 2 & 2\\[3pt]
Channel Multiplier & 1,2,3,5 & 1,2,2,4 & 1,2,3,5 & 1,2,4,6 \\[3pt]
BigGAN~\cite{biggan} up/downsampling & \xmark & \xmark & \xmark & \cmark \\[3pt]
activation rescaling~\cite{DBLP:journals/corr/abs-2011-13456, stylegan1, stylegan2} & \xmark & \xmark & \xmark & \cmark \\[3pt]
Number of Heads & 32 & 32 & 32 & 32 \\[3pt]
Batch Size & 1240 & 640 & 56 & 56 \\[3pt]
Iterations & 112K & 240K  & subset dependent$^{\mathsection}$ & subset dependent$^{\mathsection}$ \\[3pt]
Learning Rate& $\text{1.0e-4}$ &$\text{1.0e-4}$ & $\text{1.0e-4}$ & $\text{1.0e-4}$  \\[3pt]
\midrule
Conditioning & CA & CA & CA & -  \\[3pt]
CA/SA-resolutions & 32, 16, 8 & 16, 8, 4  & 32, 16, 8 & 32, 16, 8 \\[3pt]
Embedding Dimension &  512&512&512 ($\phi = \phi_{\text{\tiny CLIP}}$)/1024 ($\phi = \phi_{\text{\tiny VQGAN}}$)& - \\[3pt]
Transformers Depth & 1 & 1 & 1 & - \\[3pt]
\bottomrule
\end{tabular}

\end{adjustbox}
\vspace{1em}
\caption{\label{tab:hyperparams} Hyperparameters for the diffusion based models presented in this work.$^{*}$: All qualitative examples in this work and the numbers presented in Tab.~\ref{tab:uin_metrics} are generated with this model;$^{\dagger}$: The models trained for the $k_{\text{\tiny train}}$ experiments in Sec.~\ref{subsec:unconditional} are all trained with these hyperparameters;$^{\ddagger}$: The various semi- and fully-parametric models referred to in Sec.~\ref{subsec:dataset_complexity} are trained with these hyperparameters; $^{\mathsection}$: All models were trained until convergence.}
\end{table}
}

\newcommand{\hyperparamsRARM}{
\begin{table}[thbp]
\centering
\setlength{\tabcolsep}{5pt}

\begin{adjustbox}{max width=\linewidth}
\footnotesize
\setlength{\tabcolsep}{5pt}
\begin{tabular}{lcc}
    \toprule
                      & \armodel                  & baseline ARM              \\ [3pt]
    \midrule
    Dataset           & ImageNet-Subsets          & ImageNet-Subsets          \\ [3pt]
    Image size        & $256 \times 256 \times 3$ & $256 \times 256 \times 3$ \\ [3pt]
    Z-shape           & $16 \times 16 \times 256$ & $16 \times 16 \times 256$ \\ [3pt]
    \#Codes           & \num{16384}               & \num{16384}               \\ [3pt]
    Model Size        & 231 M                     & 265 M                     \\ [3pt]
    \#Heads           & 12                        & 14                        \\ [3pt]
    Channel per Head  & 64                        & 64                        \\ [3pt]
    Depth             & 18                        & 18                        \\ [3pt]
    Batch Size        & 100                       & 100                       \\ [3pt]
    Iterations        & subset dependent          & subset dependent          \\ [3pt]
    Learning rate     & \num{5.0e-04}             & \num{5.0e-04}             \\ [3pt]
    \midrule
    Conditioning      & CA                        & -                         \\ [3pt]
    Context Dimension & 512                       & -                         \\ [3pt]
    \bottomrule
\end{tabular}
\end{adjustbox}
\vspace{1em}
\caption{\label{tab:hyperparamsRARM} Hyperparameters for the autoregressive models used in this work. Qualitative examples and quantitative results stem from different models as described in the corresponding section. All models were trained until convergence.}
\end{table}
}

\newcommand{\imagenetsubsets}{
\begin{wraptable}{r}{.35\textwidth}
\centering
\vspace{-1em}
\begin{footnotesize}
\begin{tabular}{l@{\hspace{3pt}}c@{\hspace{3pt}}c}
\toprule
\textbf{Dataset} & class labels & $N$  \\[3pt]
\midrule
IN-dogs & 151-280 & 163K \\[3pt]
IN-mammals  & 147-388 & 309K\\[3pt]
IN-animals & 0-397 & 511K \\[3pt]
\bottomrule
\end{tabular}
\end{footnotesize}
\caption{\label{tab:in_subsets} Statistics for the ImageNet subsets used in the analysis on dataset complexity in Sec.~\ref{subsec:dataset_complexity} and Fig.~\ref{fig:dataset_complexity}.\vspace{-1em}}
\end{wraptable}
}

\providecommand{\imwidth}{}

\providecommand{\impath}[1]{}
\providecommand{\impaths}[1]{}
\providecommand{\impatha}[1]{}
\providecommand{\impathb}[1]{}
\providecommand{\impathc}[1]{}
\providecommand{\impathd}[1]{}
\providecommand{\impathe}[1]{}
\providecommand{\impathf}[1]{}
\providecommand{\impathg}[1]{}
\providecommand{\impathh}[1]{}
\providecommand{\impathi}[1]{}

\newcommand{\databasecomp}{
\begin{figure}[htbp]
\renewcommand{\imwidth}{0.49\textwidth}
\renewcommand{\impath}[1]{img/stylization/random_samples/##1}
\begin{tabular}{cc}
\toprule
$\mathcal{D}_{\text{\tiny train}}$ & $\mathcal{D}_{\text{\tiny style}}$ \\
\midrule
\includegraphics[width=\imwidth]{\impath{orig_five_1}} &
\includegraphics[width=\imwidth]{\impath{five_1}} \\

\includegraphics[width=\imwidth]{\impath{orig_five_2}} &
\includegraphics[width=\imwidth]{\impath{five_2}} \\

\includegraphics[width=\imwidth]{\impath{orig_five_3}} &
\includegraphics[width=\imwidth]{\impath{five_3}} \\

\includegraphics[width=\imwidth]{\impath{orig_five_4}} &
\includegraphics[width=\imwidth]{\impath{five_4}} \\

\includegraphics[width=\imwidth]{\impath{orig_five_5}} &
\includegraphics[width=\imwidth]{\impath{five_5}} \\

\includegraphics[width=\imwidth]{\impath{orig_five_6}} &
\includegraphics[width=\imwidth]{\impath{five_6}} \\

\includegraphics[width=\imwidth]{\impath{orig_five_7}} &
\includegraphics[width=\imwidth]{\impath{five_7}} \\

\includegraphics[width=\imwidth]{\impath{orig_five_8}} &
\includegraphics[width=\imwidth]{\impath{five_8}} \\

\includegraphics[width=\imwidth]{\impath{orig_five_9}} &
\includegraphics[width=\imwidth]{\impath{five_9}} \\

\includegraphics[width=\imwidth]{\impath{orig_five_10}} &
\includegraphics[width=\imwidth]{\impath{five_10}} \\

\includegraphics[width=\imwidth]{\impath{orig_five_11}} &
\includegraphics[width=\imwidth]{\impath{five_11}} \\

\includegraphics[width=\imwidth]{\impath{orig_five_12}} &
\includegraphics[width=\imwidth]{\impath{five_12}} \\

\bottomrule
\end{tabular}
\caption{\label{fig:random_styles_comp} Comparing random unconditional samples when replacing the train database $\mathcal{D}_{\text{\tiny train}}$ with a new database $\mathcal{D}_{\text{\tiny style}}$ consisting of the entire image corpus of WikiArt~\cite{wikiart}. Images were generated with classifier-free scale $s = 2.0$ and 100 DDIM steps.}
\end{figure}
}

\newcommand{\moreccsamples}{
\begin{figure}[htbp]
\renewcommand{\imwidth}{0.15\textwidth}
\renewcommand{\impatha}[1]{img/class_conditional/supplement/fire_salamander/##1}
\renewcommand{\impathb}[1]{img/class_conditional/supplement/bison/##1}
\renewcommand{\impathc}[1]{img/class_conditional/supplement/german_shepperd/##1}
\renewcommand{\impathd}[1]{img/class_conditional/supplement/red_fox/##1}
\renewcommand{\impathe}[1]{img/class_conditional/supplement/gorilla/##1}
\renewcommand{\impathf}[1]{img/class_conditional/supplement/pickelhaube/##1}
\renewcommand{\impathg}[1]{img/class_conditional/supplement/limo/##1}
\renewcommand{\impathh}[1]{img/class_conditional/supplement/organ/##1}
\renewcommand{\impathi}[1]{img/class_conditional/supplement/wand_clock/##1}
\setlength{\tabcolsep}{0pt}
\centering
\begin{tabular}{c@{\hspace{2pt}}cccccc}
\toprule
\parbox[c]{.1\textwidth}{\vspace{-5em}\shortstack{\scriptsize \emph{Fire} \\ \scriptsize \emph{salamander}}} &
\includegraphics[width=\imwidth]{\impatha{1}} &
\includegraphics[width=\imwidth]{\impatha{2}} &
\includegraphics[width=\imwidth]{\impatha{3}} &
\includegraphics[width=\imwidth]{\impatha{4}} &
\includegraphics[width=\imwidth]{\impatha{5}} &
\includegraphics[width=\imwidth]{\impatha{6}} \\

\midrule

\parbox[c]{.05\textwidth}{\vspace{-5em}\scriptsize \emph{Bison}} &
\includegraphics[width=\imwidth]{\impathb{1}} &
\includegraphics[width=\imwidth]{\impathb{2}} &
\includegraphics[width=\imwidth]{\impathb{3}} &
\includegraphics[width=\imwidth]{\impathb{4}} &
\includegraphics[width=\imwidth]{\impathb{5}} &
\includegraphics[width=\imwidth]{\impathb{6}} \\

\midrule

\parbox[c]{.08\textwidth}{\vspace{-5em}\shortstack{\scriptsize \emph{German} \\ \scriptsize \emph{shepherd}}} &
\includegraphics[width=\imwidth]{\impathc{1}} &
\includegraphics[width=\imwidth]{\impathc{2}} &
\includegraphics[width=\imwidth]{\impathc{3}} &
\includegraphics[width=\imwidth]{\impathc{4}} &
\includegraphics[width=\imwidth]{\impathc{5}} &
\includegraphics[width=\imwidth]{\impathc{6}} \\
\midrule

\parbox[c]{.05\textwidth}{\vspace{-5em}\shortstack{\scriptsize \emph{Red} \\ \scriptsize \emph{fox}}} &
\includegraphics[width=\imwidth]{\impathd{1}} &
\includegraphics[width=\imwidth]{\impathd{2}} &
\includegraphics[width=\imwidth]{\impathd{3}} &
\includegraphics[width=\imwidth]{\impathd{4}} &
\includegraphics[width=\imwidth]{\impathd{5}} &
\includegraphics[width=\imwidth]{\impathd{6}} \\

\midrule

\parbox[c]{.07\textwidth}{\vspace{-5em}\scriptsize\emph{Gorilla}} &
\includegraphics[width=\imwidth]{\impathe{1}} &
\includegraphics[width=\imwidth]{\impathe{2}} &
\includegraphics[width=\imwidth]{\impathe{3}} &
\includegraphics[width=\imwidth]{\impathe{4}} &
\includegraphics[width=\imwidth]{\impathe{5}} &
\includegraphics[width=\imwidth]{\impathe{6}} \\

\midrule

\parbox[c]{.1\textwidth}{\vspace{-5em}\scriptsize\emph{Pickelhaube} }&
\includegraphics[width=\imwidth]{\impathf{1}} &
\includegraphics[width=\imwidth]{\impathf{2}} &
\includegraphics[width=\imwidth]{\impathf{3}} &
\includegraphics[width=\imwidth]{\impathf{4}} &
\includegraphics[width=\imwidth]{\impathf{5}} &
\includegraphics[width=\imwidth]{\impathf{6}} \\

\midrule

\parbox[c]{.09\textwidth}{\vspace{-5em}\scriptsize\emph{Limousine}} &
\includegraphics[width=\imwidth]{\impathg{1}} &
\includegraphics[width=\imwidth]{\impathg{2}} &
\includegraphics[width=\imwidth]{\impathg{3}} &
\includegraphics[width=\imwidth]{\impathg{4}} &
\includegraphics[width=\imwidth]{\impathg{5}} &
\includegraphics[width=\imwidth]{\impathg{6}} \\

\midrule

\parbox[c]{.07\textwidth}{\vspace{-5em}\scriptsize\emph{Organ}} &
\includegraphics[width=\imwidth]{\impathh{1}} &
\includegraphics[width=\imwidth]{\impathh{2}} &
\includegraphics[width=\imwidth]{\impathh{3}} &
\includegraphics[width=\imwidth]{\impathh{4}} &
\includegraphics[width=\imwidth]{\impathh{5}} &
\includegraphics[width=\imwidth]{\impathh{6}} \\

\midrule

\parbox[c]{.06\textwidth}{\vspace{-5em}\shortstack{\scriptsize \emph{Wand} \\ \scriptsize \emph{clock}}}  &
\includegraphics[width=\imwidth]{\impathi{1}} &
\includegraphics[width=\imwidth]{\impathi{2}} &
\includegraphics[width=\imwidth]{\impathi{3}} &
\includegraphics[width=\imwidth]{\impathi{4}} &
\includegraphics[width=\imwidth]{\impathi{5}} &
\includegraphics[width=\imwidth]{\impathi{6}} \\

\bottomrule

\end{tabular}
\caption{\label{fig:more_cc_samples} Additional class conditional samples obtained via the conditioning method presented in Sec.~\ref{subsec:model_inference}. Samples are generated with classifier-free scale $s = 2.0$ and 100 DDIM steps.}
\end{figure}
}

\newcommand{\topmvisual}{
\begin{figure}[htbp]
\renewcommand{\imwidth}{0.163\textwidth}
\renewcommand{\impath}[1]{img/memsize_influence/##1}
\setlength{\tabcolsep}{2pt}
\centering
\begin{tabular}{cccccc}
\toprule
\footnotesize Single Example & \footnotesize $m = 10^{-5}$ & \footnotesize $m = 10^{-4}$ & \footnotesize $m = 10^{-3}$ & \footnotesize $m = 10^{-2}$ & \footnotesize $m = 10^{-1}$ \\
\toprule
\includegraphics[width=\imwidth]{\impath{memsize_1_complete}} &
\includegraphics[width=\imwidth]{\impath{memsize_-5_complete}} &
\includegraphics[width=\imwidth]{\impath{memsize_-4_complete}} &
\includegraphics[width=\imwidth]{\impath{memsize_-3_complete}} &
\includegraphics[width=\imwidth]{\impath{memsize_-2_complete}} &
\includegraphics[width=\imwidth]{\impath{memsize_-1_complete}} \\

\bottomrule
\end{tabular}
\caption{\label{fig:topm_visual} Visual examples on the quality-diversity trade off obtained by \emph{top-m sampling}. For heavily truncated $p_{\mathcal{D}}(\tilde{x})$ we obtain extremely low sample diversity as visualized in the examples on the left part. Increasing $m$ results in more diversity but lower sample fidelity (right part). All images generated with guidance scale $s=1.5$ and 100 DDIM steps.}
\end{figure}
}

\newcommand{\rarmtopmvisual}{
\begin{figure}[htbp]
\renewcommand{\imwidth}{0.163\textwidth}
\renewcommand{\impath}[1]{img/transformer_additional/top-m/##1}
\setlength{\tabcolsep}{2pt}
\centering
\begin{tabular}{cccccc}
\toprule
\footnotesize Single Example & \footnotesize $m = 10^{-5}$ & \footnotesize $m = 10^{-4}$ & \footnotesize $m = 10^{-3}$ & \footnotesize $m = 10^{-2}$ & \footnotesize $m = 10^{-1}$ \\
\toprule
\includegraphics[width=\imwidth]{\impath{memsize_1_complete}} &
\includegraphics[width=\imwidth]{\impath{memsize_-5_complete}} &
\includegraphics[width=\imwidth]{\impath{memsize_-4_complete}} &
\includegraphics[width=\imwidth]{\impath{memsize_-3_complete}} &
\includegraphics[width=\imwidth]{\impath{memsize_-2_complete}} &
\includegraphics[width=\imwidth]{\impath{memsize_-1_complete}} \\
\bottomrule
\end{tabular}
\caption{\label{fig:topm_visual_rarm} Visual examples on the quality-diversity trade off obtained by \emph{top-m sampling} using our \armodel trained on IN-animals. For heavily truncated $p_{\mathcal{D}}(\tilde{x})$ we obtain extremely low sample diversity as visualized in the examples on the left part. Increasing $m$ results in more diversity but lower sample fidelity (right part). All images generated with guidance scale $s=2.0$ and generated with top-\(k=4096\). Note that this model is trained on the Animals subset of ImageNet. Therefore, the proposal distribution $p_{\mathcal{D}}(\tilde{x})$ differs from that of the shown results for \diffusionmodel in Fig.~\ref{fig:topm_visual}, which is trained on the entire ImageNet dataset. This is the reason for the different classes of dogs for the leftmost column of this Figure compared to Fig.~\ref{fig:topm_visual}.}
\end{figure}
}

\newcommand{\patchsizefig}{
\begin{wrapfigure}{r}{.34\textwidth}
\vspace{-1em}
\renewcommand{\impath}[1]{img/##1}
\includegraphics[width=.34\textwidth]{\impath{database_patchsize2}}
\caption{\label{fig:patchsize} Effect of patch size of images in the retrieval database.
\vspace{-2em}}
\end{wrapfigure}
}

\newcommand{\nnreps}{
\begin{figure}[htbp]%{r}{.34\textwidth}
\vspace{-1em}
\centering
\renewcommand{\impath}[1]{img/##1}
\includegraphics[width=.4\textwidth]{\impath{nn_rep}}

\caption{\label{fig:nn_rep} Performance of \diffusionmodel with different nearest neighbor representations.
\vspace{-2em}}

\end{figure}
}

\newcommand{\retroguidingsamples}{
\begin{figure}[htbp]
\renewcommand{\imwidth}{0.19\textwidth}
\renewcommand{\impath}[1]{img/retrieval_guidance/new/##1}
\setlength{\tabcolsep}{.5pt}
\centering
\begin{tabular}{c@{\hspace{1pt}}c@{\hspace{2pt}}cccc}
\toprule
\multicolumn{2}{c}{\footnotesize Neighbors }& \footnotesize $s=1$ & \footnotesize $s=2$ & \footnotesize $s=3$ & \footnotesize $s=4$ \\
\toprule
\shortstack{\includegraphics[width=.095\textwidth]{\impath{nns1_1-sample0}} \\[-3pt]
            \includegraphics[width=.095\textwidth]{\impath{nns1_2-sample0}}} &
\shortstack{\includegraphics[width=.095\textwidth]{\impath{nns2_1-sample0}} \\[-3pt]
            \includegraphics[width=.095\textwidth]{\impath{nns2_2-sample0}}} &
\includegraphics[width=\imwidth]{\impath{s1-samples-sample0}} &
\includegraphics[width=\imwidth]{\impath{s2-samples-sample0}} &
\includegraphics[width=\imwidth]{\impath{s3-samples-sample0}} &
\includegraphics[width=\imwidth]{\impath{s4-samples-sample0}} \\

\shortstack{\includegraphics[width=.095\textwidth]{\impath{nns1_1-sample1}} \\[-3pt]
            \includegraphics[width=.095\textwidth]{\impath{nns1_2-sample1}}} &
\shortstack{\includegraphics[width=.095\textwidth]{\impath{nns2_1-sample1}} \\[-3pt]
            \includegraphics[width=.095\textwidth]{\impath{nns2_2-sample1}}} &
\includegraphics[width=\imwidth]{\impath{s1-samples-sample1}} &
\includegraphics[width=\imwidth]{\impath{s2-samples-sample1}} &
\includegraphics[width=\imwidth]{\impath{s3-samples-sample1}} &
\includegraphics[width=\imwidth]{\impath{s4-samples-sample1}} \\

\shortstack{\includegraphics[width=.095\textwidth]{\impath{nns1_1-sample2}} \\[-3pt]
            \includegraphics[width=.095\textwidth]{\impath{nns1_2-sample2}}} &
\shortstack{\includegraphics[width=.095\textwidth]{\impath{nns2_1-sample2}} \\[-3pt]
            \includegraphics[width=.095\textwidth]{\impath{nns2_2-sample2}}} &
\includegraphics[width=\imwidth]{\impath{s1-samples-sample2}} &
\includegraphics[width=\imwidth]{\impath{s2-samples-sample2}} &
\includegraphics[width=\imwidth]{\impath{s3-samples-sample2}} &
\includegraphics[width=\imwidth]{\impath{s4-samples-sample2}} \\

\shortstack{\includegraphics[width=.095\textwidth]{\impath{nns1_1-sample3}} \\[-2.8pt]
            \includegraphics[width=.095\textwidth]{\impath{nns1_2-sample3}}} &
\shortstack{\includegraphics[width=.095\textwidth]{\impath{nns2_1-sample3}} \\[-2.8pt]
            \includegraphics[width=.095\textwidth]{\impath{nns2_2-sample3}}} &
\includegraphics[width=\imwidth]{\impath{s1-samples-sample3}} &
\includegraphics[width=\imwidth]{\impath{s2-samples-sample3}} &
\includegraphics[width=\imwidth]{\impath{s3-samples-sample3}} &
\includegraphics[width=\imwidth]{\impath{s4-samples-sample3}} \\

\shortstack{\includegraphics[width=.095\textwidth]{\impath{nns1_1-sample4}} \\[-2.8pt]
            \includegraphics[width=.095\textwidth]{\impath{nns1_2-sample4}}} &
\shortstack{\includegraphics[width=.095\textwidth]{\impath{nns2_1-sample4}} \\[-2.8pt]
            \includegraphics[width=.095\textwidth]{\impath{nns2_2-sample4}}} &
\includegraphics[width=\imwidth]{\impath{s1-samples-sample4}} &
\includegraphics[width=\imwidth]{\impath{s2-samples-sample4}} &
\includegraphics[width=\imwidth]{\impath{s3-samples-sample4}} &
\includegraphics[width=\imwidth]{\impath{s4-samples-sample4}} \\
\bottomrule
\end{tabular}
\caption{\label{fig:retro_guiding_samples} Visualizing the effects of retrieval based classifier free guidance. All images generated with fixed random seed, $m=0.1$ and 100 DDIM steps.}
\end{figure}
}

\newcommand{\rarmguidingsamples}{
\begin{figure}[htbp]
\renewcommand{\imwidth}{0.095\textwidth}
\renewcommand{\impath}[1]{img/transformer_additional/guidance/##1}
\setlength{\tabcolsep}{.5pt}
\centering
\begin{tabular}{c@{\hspace{1pt}}c@{\hspace{2pt}}cccc}
\toprule
\multicolumn{2}{c}{\footnotesize Neighbors }& \footnotesize $s=1$ & \footnotesize $s=2$ & \footnotesize $s=3$ & \footnotesize $s=4$ \\
\toprule
\includegraphics[width=\imwidth]{\impath{2022-05-25-18-35-39-batched_nns-run0-sample0}} &
\includegraphics[width=\imwidth]{\impath{2022-05-25-18-35-39-batched_nns-run0-sample1}}
&
\includegraphics[width=\imwidth]{\impath{2022-05-25-18-35-39-samples_with_sampled_nns-run0-sample0}}\hspace{-3pt}
\includegraphics[width=\imwidth]{\impath{2022-05-25-18-35-39-samples_with_sampled_nns-run0-sample1}}
&
\includegraphics[width=\imwidth]{\impath{2022-05-25-18-35-39-samples_with_sampled_nns-run1-sample0}}\hspace{-3pt}
\includegraphics[width=\imwidth]{\impath{2022-05-25-18-35-39-samples_with_sampled_nns-run1-sample1}}
&
\includegraphics[width=\imwidth]{\impath{2022-05-25-18-35-39-samples_with_sampled_nns-run2-sample0}}\hspace{-3pt}
\includegraphics[width=\imwidth]{\impath{2022-05-25-18-35-39-samples_with_sampled_nns-run2-sample1}}
&
\includegraphics[width=\imwidth]{\impath{2022-05-25-18-35-39-samples_with_sampled_nns-run3-sample0}}\hspace{-3pt}
\includegraphics[width=\imwidth]{\impath{2022-05-25-18-35-39-samples_with_sampled_nns-run3-sample1}}
    \\
\includegraphics[width=\imwidth]{\impath{2022-05-25-18-35-39-batched_nns-run0-sample2}} &
\includegraphics[width=\imwidth]{\impath{2022-05-25-18-35-39-batched_nns-run0-sample3}}
&
\includegraphics[width=\imwidth]{\impath{2022-05-25-18-35-39-samples_with_sampled_nns-run0-sample2}}\hspace{-3pt}
\includegraphics[width=\imwidth]{\impath{2022-05-25-18-35-39-samples_with_sampled_nns-run0-sample3}}
&
\includegraphics[width=\imwidth]{\impath{2022-05-25-18-35-39-samples_with_sampled_nns-run1-sample2}}\hspace{-3pt}
\includegraphics[width=\imwidth]{\impath{2022-05-25-18-35-39-samples_with_sampled_nns-run1-sample3}}
&
\includegraphics[width=\imwidth]{\impath{2022-05-25-18-35-39-samples_with_sampled_nns-run2-sample2}}\hspace{-3pt}
\includegraphics[width=\imwidth]{\impath{2022-05-25-18-35-39-samples_with_sampled_nns-run2-sample3}}
&
\includegraphics[width=\imwidth]{\impath{2022-05-25-18-35-39-samples_with_sampled_nns-run3-sample2}}\hspace{-3pt}
\includegraphics[width=\imwidth]{\impath{2022-05-25-18-35-39-samples_with_sampled_nns-run3-sample3}}
    \\
\includegraphics[width=\imwidth]{\impath{2022-05-25-18-35-39-batched_nns-run0-sample4}} &
\includegraphics[width=\imwidth]{\impath{2022-05-25-18-35-39-batched_nns-run0-sample5}}
&
\includegraphics[width=\imwidth]{\impath{2022-05-25-18-35-39-samples_with_sampled_nns-run0-sample4}}\hspace{-3pt}
\includegraphics[width=\imwidth]{\impath{2022-05-25-18-35-39-samples_with_sampled_nns-run0-sample5}}
&
\includegraphics[width=\imwidth]{\impath{2022-05-25-18-35-39-samples_with_sampled_nns-run1-sample4}}\hspace{-3pt}
\includegraphics[width=\imwidth]{\impath{2022-05-25-18-35-39-samples_with_sampled_nns-run1-sample5}}
&
\includegraphics[width=\imwidth]{\impath{2022-05-25-18-35-39-samples_with_sampled_nns-run2-sample4}}\hspace{-3pt}
\includegraphics[width=\imwidth]{\impath{2022-05-25-18-35-39-samples_with_sampled_nns-run2-sample5}}
&
\includegraphics[width=\imwidth]{\impath{2022-05-25-18-35-39-samples_with_sampled_nns-run3-sample4}}\hspace{-3pt}
\includegraphics[width=\imwidth]{\impath{2022-05-25-18-35-39-samples_with_sampled_nns-run3-sample5}}
    \\
\includegraphics[width=\imwidth]{\impath{2022-05-25-18-35-39-batched_nns-run0-sample6}} &
\includegraphics[width=\imwidth]{\impath{2022-05-25-18-35-39-batched_nns-run0-sample7}}
&
\includegraphics[width=\imwidth]{\impath{2022-05-25-18-35-39-samples_with_sampled_nns-run0-sample6}}\hspace{-3pt}
\includegraphics[width=\imwidth]{\impath{2022-05-25-18-35-39-samples_with_sampled_nns-run0-sample7}}
&
\includegraphics[width=\imwidth]{\impath{2022-05-25-18-35-39-samples_with_sampled_nns-run1-sample6}}\hspace{-3pt}
\includegraphics[width=\imwidth]{\impath{2022-05-25-18-35-39-samples_with_sampled_nns-run1-sample7}}
&
\includegraphics[width=\imwidth]{\impath{2022-05-25-18-35-39-samples_with_sampled_nns-run2-sample6}}\hspace{-3pt}
\includegraphics[width=\imwidth]{\impath{2022-05-25-18-35-39-samples_with_sampled_nns-run2-sample7}}
&
\includegraphics[width=\imwidth]{\impath{2022-05-25-18-35-39-samples_with_sampled_nns-run3-sample6}}\hspace{-3pt}
\includegraphics[width=\imwidth]{\impath{2022-05-25-18-35-39-samples_with_sampled_nns-run3-sample7}}
    \\
\includegraphics[width=\imwidth]{\impath{2022-05-25-18-35-39-batched_nns-run0-sample8}} &
\includegraphics[width=\imwidth]{\impath{2022-05-25-18-35-39-batched_nns-run0-sample9}}
&
\includegraphics[width=\imwidth]{\impath{2022-05-25-18-35-39-samples_with_sampled_nns-run0-sample8}}\hspace{-3pt}
\includegraphics[width=\imwidth]{\impath{2022-05-25-18-35-39-samples_with_sampled_nns-run0-sample9}}
&
\includegraphics[width=\imwidth]{\impath{2022-05-25-18-35-39-samples_with_sampled_nns-run1-sample8}}\hspace{-3pt}
\includegraphics[width=\imwidth]{\impath{2022-05-25-18-35-39-samples_with_sampled_nns-run1-sample9}}
&
\includegraphics[width=\imwidth]{\impath{2022-05-25-18-35-39-samples_with_sampled_nns-run2-sample8}}\hspace{-3pt}
\includegraphics[width=\imwidth]{\impath{2022-05-25-18-35-39-samples_with_sampled_nns-run2-sample9}}
&
\includegraphics[width=\imwidth]{\impath{2022-05-25-18-35-39-samples_with_sampled_nns-run3-sample8}}\hspace{-3pt}
\includegraphics[width=\imwidth]{\impath{2022-05-25-18-35-39-samples_with_sampled_nns-run3-sample9}}
    \\
\bottomrule
\end{tabular}
\caption{\label{fig:rarm_guiding_samples} Visualizing the effects of retrieval based classifier-free guidance for the \armodel trained on IN-animals. All images generated with fixed random seed, $m=0.01$ and top-\(k=4096\).}
\end{figure}
}

\newcommand{\suppstylizer}{
\begin{figure}[htbp]
\renewcommand{\imwidth}{0.125\textwidth}
\renewcommand{\impath}[1]{img/stylization/supplement/##1}
\setlength{\tabcolsep}{2pt}
\centering
\begin{adjustbox}{max width=\textwidth}
\begin{tabular}{cccccccc}
\toprule
\shortstack{\tiny\emph{'The lion, } \\ \tiny\emph{ king of beasts.'}} & \shortstack{\tiny\emph{'A red sun} \\ \tiny\emph{ is drowning.'}} & \shortstack{\tiny\emph{'A mighty,} \\ \tiny\emph{ old horse.'}} & \shortstack{\tiny\emph{'A marriage} \\ \tiny\emph{in the forest.'}}  &\shortstack{\tiny\emph{'A phoenix} \\ \tiny\emph{in the sky.'}} &  \shortstack{\tiny\emph{'An assembly} \\ \tiny\emph{ of aristocrats.'}}&  \shortstack{\tiny\emph{'Shadows on} \\ \tiny\emph{a wall.'}}& \shortstack{\tiny\emph{'A forest} \\ \tiny\emph{ in fall.'}} \\
\toprule
\includegraphics[width=\imwidth]{\impath{lion_1}} &
\includegraphics[width=\imwidth]{\impath{a_red_sun_1}} &
\includegraphics[width=\imwidth]{\impath{mighty_old_horse_1}} &
\includegraphics[width=\imwidth]{\impath{marriage_single_1}} &
\includegraphics[width=\imwidth]{\impath{phoenix_in_the_sky_4}} &
\includegraphics[width=\imwidth]{\impath{An_assembly_of_the_aristocrats_1}} &
\includegraphics[width=\imwidth]{\impath{shadows_on_the_wall_single_1}} &
\includegraphics[width=\imwidth]{\impath{forest_in_fall_1}} \\[-3pt]

\includegraphics[width=\imwidth]{\impath{lion_2}} &
\includegraphics[width=\imwidth]{\impath{a_red_sun_2}} &
\includegraphics[width=\imwidth]{\impath{mighty_old_horse_2}} &
\includegraphics[width=\imwidth]{\impath{marriage_single_2}} &
\includegraphics[width=\imwidth]{\impath{phoenix_in_the_sky_2}} &
\includegraphics[width=\imwidth]{\impath{An_assembly_of_the_aristocrats_2}} &
\includegraphics[width=\imwidth]{\impath{shadows_on_the_wall_single_2}} &
\includegraphics[width=\imwidth]{\impath{forest_in_fall_2}} \\

\includegraphics[width=\imwidth]{\impath{lion_3}} &
\includegraphics[width=\imwidth]{\impath{a_red_sun_3}} &
\includegraphics[width=\imwidth]{\impath{mighty_old_horse_3}} &
\includegraphics[width=\imwidth]{\impath{marriage_single_3}} &
\includegraphics[width=\imwidth]{\impath{phoenix_in_the_sky_3}} &
\includegraphics[width=\imwidth]{\impath{An_assembly_of_the_aristocrats_3}} &
\includegraphics[width=\imwidth]{\impath{shadows_on_the_wall_single_3}} &
\includegraphics[width=\imwidth]{\impath{forest_in_fall_3}} \\

\includegraphics[width=\imwidth]{\impath{lion_4}} &
\includegraphics[width=\imwidth]{\impath{a_red_sun_4}} &
\includegraphics[width=\imwidth]{\impath{mighty_old_horse_4}} &
\includegraphics[width=\imwidth]{\impath{marriage_single_4}} &
\includegraphics[width=\imwidth]{\impath{phoenix_in_the_sky_1}} &
\includegraphics[width=\imwidth]{\impath{An_assembly_of_the_aristocrats_5}} &
\includegraphics[width=\imwidth]{\impath{shadows_on_the_wall_single_4}} &
\includegraphics[width=\imwidth]{\impath{forest_in_fall_4}} \\
\bottomrule

\end{tabular}
\end{adjustbox}
\caption{\label{fig:stylizer_supp} Additional samples for zero-shot text-guided stylization with our ImageNet \diffusionmodel as in Fig.~\ref{fig:stylize}. Samples are generated with classifier-free scale $s = 2.5$ and 100 DDIM steps.}
\end{figure}
}

\newcommand{\uinrandomsamplestwo}{
\begin{figure}[t]
\renewcommand{\imwidth}{\linewidth}
\renewcommand{\impath}[1]{img/##1}
\centering
\begin{tabular}{c}
\toprule
\includegraphics[width=\linewidth]{\impath{uin_row1}}\\[-3pt]
\includegraphics[width=\linewidth]{\impath{uin_row2}}\\[-3pt]
\includegraphics[width=\linewidth]{\impath{uin_row3}}\\[-3pt]
\includegraphics[width=\linewidth]{\impath{uin_row4}}\\[-3pt]
\includegraphics[width=\linewidth]{\impath{uin_row5}}\\[-3pt]
\includegraphics[width=\linewidth]{\impath{uin_row6}}\\[-3pt]
\includegraphics[width=\linewidth]{\impath{uin_row7}}\\[-3pt]
\includegraphics[width=\linewidth]{\impath{uin_row8}}\\[-3pt]
\includegraphics[width=\linewidth]{\impath{uin_row9}}\\[-3pt]
\includegraphics[width=\linewidth]{\impath{uin_row10}}\\[-3pt]
\includegraphics[width=\linewidth]{\impath{uin_row11}}\\[-3pt]
\includegraphics[width=\linewidth]{\impath{uin_row12}}\\[-3pt]

\bottomrule
\end{tabular}
\caption{\label{fig:uin_rsamples} Random samples from our \diffusionmodel, with $m=0.01$ and classifier-free guidance with $s = 2.0$. Samples were generated with 100 DDIM steps.}
\end{figure}
}

\newcommand{\rarmrandomsamplestwo}{
\begin{figure}[t]
\renewcommand{\imwidth}{\linewidth}
\renewcommand{\impath}[1]{img/transformer_additional/##1}
\centering
\begin{tabular}{c}
\toprule
ImageNet-Dogs \\
\midrule
\includegraphics[width=\linewidth]{\impath{dogs_row1}}\\[-3pt]
\includegraphics[width=\linewidth]{\impath{dogs_row2}}\\[-3pt]
\includegraphics[width=\linewidth]{\impath{dogs_row3}}\\[-3pt]
\includegraphics[width=\linewidth]{\impath{dogs_row4}}\\[2pt]
\midrule
ImageNet-Mammals \\
\midrule
\includegraphics[width=\linewidth]{\impath{mammals_row1}}\\[-3pt]
\includegraphics[width=\linewidth]{\impath{mammals_row2}}\\[-3pt]
\includegraphics[width=\linewidth]{\impath{mammals_row3}}\\[-3pt]
\includegraphics[width=\linewidth]{\impath{mammals_row4}}\\[2pt]
\midrule
ImageNet-Animals \\
\midrule
\includegraphics[width=\linewidth]{\impath{animals_row1}}\\[-3pt]
\includegraphics[width=\linewidth]{\impath{animals_row2}}\\[-3pt]
\includegraphics[width=\linewidth]{\impath{animals_row3}}\\[-3pt]
\includegraphics[width=\linewidth]{\impath{animals_row4}}\\[-3pt]
\bottomrule
\end{tabular}
\caption{\label{fig:rarm_rsamples} Random samples from our autoregressive models, with $m=0.01$ and classifier-free guidance with $s = 2.0$.
The models are trained on the dogs subset (top rows), mammals subset (middle rows), and animal subset (bottom rows).
Samples were generated with top-$k=4096$.}
\end{figure}
}

\newcommand{\cliptextquerysupp}{
\begin{figure}[htbp]
\renewcommand{\imwidth}{0.125\textwidth}
\renewcommand{\impath}[1]{img/text-guiding/##1}
\setlength{\tabcolsep}{2pt}
\centering
\begin{adjustbox}{max width=\textwidth}
\begin{tabular}{cccccccc}
\toprule
\shortstack{\tiny\emph{'A turtle with a} \\ \tiny\emph{ shell made of gold.'}} & \shortstack{\tiny\emph{'A turtle with a} \\ \tiny\emph{ shell made of silver.'}} & \shortstack{\tiny\emph{'A cyborg koala} \\ \tiny\emph{ wearing an armor.'}} & \shortstack{\tiny\emph{'A brown bear rea-} \\ \tiny\emph{ ding a newspaper.'}}  &\shortstack{\tiny\emph{'An apple with black } \\ \tiny\emph{and white stripes.'}} &  \shortstack{\tiny\emph{'A pink elephant} \\ \tiny\emph{ in the savannah.'}}&  \shortstack{\tiny\emph{'Vector illustration of} \\ \tiny\emph{ a red tiger head.'}}& \shortstack{\tiny\emph{'A pizza} \\ \tiny\emph{ made of wood.'}} \\
\toprule
\includegraphics[width=\imwidth]{\impath{A_turtle_with_a_shell_made_of_gold_1}} &
\includegraphics[width=\imwidth]{\impath{A_turtle_with_a_shell_made_of_silver_1}} &
\includegraphics[width=\imwidth]{\impath{cyborg_koala_armour_1}} &
\includegraphics[width=\imwidth]{\impath{Brownbear_reading_a_newspaper_1}} &
\includegraphics[width=\imwidth]{\impath{apple_with_black_and_white_stripes_1}} &
\includegraphics[width=\imwidth]{\impath{pink_elephant_1}} &
\includegraphics[width=\imwidth]{\impath{vector_illustration_of_a_red_tiger_head_1}} &
\includegraphics[width=\imwidth]{\impath{pizza_made_of_wood_1}} \\[-3pt]

\includegraphics[width=\imwidth]{\impath{A_turtle_with_a_shell_made_of_gold_2}} &
\includegraphics[width=\imwidth]{\impath{A_turtle_with_a_shell_made_of_silver_2}} &
\includegraphics[width=\imwidth]{\impath{cyborg_koala_armour_2}} &
\includegraphics[width=\imwidth]{\impath{Brownbear_reading_a_newspaper_2}} &
\includegraphics[width=\imwidth]{\impath{apple_with_black_and_white_stripes_2}} &
\includegraphics[width=\imwidth]{\impath{pink_elephant_4}} &
\includegraphics[width=\imwidth]{\impath{vector_illustration_of_a_red_tiger_head_2}} &
\includegraphics[width=\imwidth]{\impath{pizza_made_of_wood_2}} \\

\includegraphics[width=\imwidth]{\impath{A_turtle_with_a_shell_made_of_gold_3}} &
\includegraphics[width=\imwidth]{\impath{A_turtle_with_a_shell_made_of_silver_3}} &
\includegraphics[width=\imwidth]{\impath{cyborg_koala_armour_3}} &
\includegraphics[width=\imwidth]{\impath{Brownbear_reading_a_newspaper_3}} &
\includegraphics[width=\imwidth]{\impath{apple_with_black_and_white_stripes_3}} &
\includegraphics[width=\imwidth]{\impath{pink_elephant_5}} &
\includegraphics[width=\imwidth]{\impath{vector_illustration_of_a_red_tiger_head_3}} &
\includegraphics[width=\imwidth]{\impath{pizza_made_of_wood_5}} \\

\includegraphics[width=\imwidth]{\impath{A_turtle_with_a_shell_made_of_gold_4}} &
\includegraphics[width=\imwidth]{\impath{A_turtle_with_a_shell_made_of_silver_4}} &
\includegraphics[width=\imwidth]{\impath{cyborg_koala_armour_4}} &
\includegraphics[width=\imwidth]{\impath{Brownbear_reading_a_newspaper_4}} &
\includegraphics[width=\imwidth]{\impath{apple_with_black_and_white_stripes_4}} &
\includegraphics[width=\imwidth]{\impath{pink_elephant_6}} &
\includegraphics[width=\imwidth]{\impath{vector_illustration_of_a_red_tiger_head_4}} &
\includegraphics[width=\imwidth]{\impath{pizza_made_of_wood_4}} \\
\bottomrule

\end{tabular}
\end{adjustbox}
\caption{\label{fig:clipretrosupp} Additional zero-shot text to image samples from our model as in Fig.~\ref{fig:clipretro}. Samples are generated with classifier-free scale $s = 2.5$ and 100 DDIM steps.}
\end{figure}
}

\newcommand{\icganclassifier}{
\renewcommand{\impath}[1]{img/imgur_update/##1}
\begin{wrapfigure}{r}{.4\textwidth}
\vspace{-1.5em}
\includegraphics[width=\linewidth]{\impath{classifier_cropped}}
\caption{\label{fig:ic_classifier} Evaluating accuracy of a binary classifier trained distinguishing between WikiArt and ImageNet on generated samples for IC-GAN and \oidiffusion. \vspace{-1em}}
\vspace{-1em}
\end{wrapfigure}
}

\newcommand{\rarmtopmtruncationwrapped}{
\renewcommand{\imwidth}{0.2\textwidth}
\renewcommand{\impath}[1]{img/topm_truncation/rarm/##1}
\begin{wrapfigure}{r}{.41\textwidth}
\vspace{-1em}
\includegraphics[width=\imwidth]{\impath{prec_rec}}
\includegraphics[width=\imwidth]{\impath{is_fid}}
\caption{\label{fig:topm_trunc_rarm} Quality-diversity trade-offs when applying top-m sampling with \armodel. \vspace{-1em}}
\end{wrapfigure}
}

\newcommand{\rarmknnwrapped}{
\begin{wrapfigure}{r}{.41\textwidth}
\vspace{-1em}
\renewcommand{\imwidth}{0.2\textwidth}
\renewcommand{\impath}[1]{img/n_nns_train/rarm/##1}
\includegraphics[width=\imwidth]{\impath{is_fid}}
\hfill
\includegraphics[width=\imwidth]{\impath{prec_rec}}
\caption{\label{fig:rarm_knn_wrapped} Effect of $k_{\text{\scriptsize train}}$ for \armodel.\vspace{-1em}}

\end{wrapfigure}
}

\newcommand{\icgangulaschsuppe}{
\renewcommand{\impatha}[1]{img/rdm_comp_icgan/sampled/oi/##1}
\renewcommand{\impathb}[1]{img/rdm_comp_icgan/sampled/wikiart/##1}
\renewcommand{\impathc}[1]{img/icgan/##1}
\renewcommand{\imwidth}{.08\textwidth}
\setlength{\tabcolsep}{0pt}
\begin{figure}
\centering
\begin{tabular}{
    c@{\hspace{0pt}}c@{\hspace{2pt}}
    c@{\hspace{0pt}}c@{\hspace{2pt}}
    c@{\hspace{0pt}}c@{\hspace{6pt}}
    c@{\hspace{0pt}}c@{\hspace{2pt}}
    c@{\hspace{0pt}}c@{\hspace{2pt}}
    c@{\hspace{0pt}}c}
\toprule
\multicolumn{6}{c}{\diffusionmodel} & \multicolumn{6}{c}{IC-GAN~\cite{ic-gan}} \\[6pt]
\multicolumn{2}{c}{$\mathcal{D}_{\text{\tiny train}}$}
    & \multicolumn{2}{c}{Pacs Cartoon}
    & \multicolumn{2}{c}{WikiArt}
& \multicolumn{2}{c}{$\mathcal{D}_{\text{\tiny train}}$}
    & \multicolumn{2}{c}{Pacs Cartoon}
    & \multicolumn{2}{c}{WikiArt}
\\ \cmidrule(lr{12pt}){1-6} \cmidrule(lr){7-12}

\includegraphics[width=\imwidth]{\impatha{sample_000043}} & \includegraphics[width=\imwidth]{\impatha{sample_000096}} & \includegraphics[width=\imwidth]{\impathc{rdm_cartoon_sample_0}} & \includegraphics[width=\imwidth]{\impathc{rdm_cartoon_sample_1}} & \includegraphics[width=\imwidth]{\impathb{sample_000000}} & \includegraphics[width=\imwidth]{\impathb{sample_000002}} & \includegraphics[width=\imwidth]{\impathc{ic_imagenet_sample_0}} & \includegraphics[width=\imwidth]{\impathc{ic_imagenet_sample_1}} & \includegraphics[width=\imwidth]{\impathc{ic_cartoon_sample_0}} & \includegraphics[width=\imwidth]{\impathc{ic_cartoon_sample_1}} & \includegraphics[width=\imwidth]{\impathc{ic_wikiart_sample_0}} & \includegraphics[width=\imwidth]{\impathc{ic_wikiart_sample_1}} \\

\includegraphics[width=\imwidth]{\impatha{sample_000093}} & \includegraphics[width=\imwidth]{\impatha{sample_000007}} & \includegraphics[width=\imwidth]{\impathc{rdm_cartoon_sample_2}} & \includegraphics[width=\imwidth]{\impathc{rdm_cartoon_sample_3}} & \includegraphics[width=\imwidth]{\impathb{sample_000017}} & \includegraphics[width=\imwidth]{\impathb{sample_000035}} & \includegraphics[width=\imwidth]{\impathc{ic_imagenet_sample_2}} & \includegraphics[width=\imwidth]{\impathc{ic_imagenet_sample_3}} & \includegraphics[width=\imwidth]{\impathc{ic_cartoon_sample_2}} & \includegraphics[width=\imwidth]{\impathc{ic_cartoon_sample_3}} & \includegraphics[width=\imwidth]{\impathc{ic_wikiart_sample_2}} & \includegraphics[width=\imwidth]{\impathc{ic_wikiart_sample_3}} \\

\includegraphics[width=\imwidth]{\impatha{sample_000012}} & \includegraphics[width=\imwidth]{\impatha{sample_000091}} & \includegraphics[width=\imwidth]{\impathc{rdm_cartoon_sample_4}} & \includegraphics[width=\imwidth]{\impathc{rdm_cartoon_sample_5}} & \includegraphics[width=\imwidth]{\impathb{sample_000057}} & \includegraphics[width=\imwidth]{\impathb{sample_000049}} & \includegraphics[width=\imwidth]{\impathc{ic_imagenet_sample_4}} & \includegraphics[width=\imwidth]{\impathc{ic_imagenet_sample_5}} & \includegraphics[width=\imwidth]{\impathc{ic_cartoon_sample_4}} & \includegraphics[width=\imwidth]{\impathc{ic_cartoon_sample_5}} & \includegraphics[width=\imwidth]{\impathc{ic_wikiart_sample_4}} & \includegraphics[width=\imwidth]{\impathc{ic_wikiart_sample_5}} \\

\includegraphics[width=\imwidth]{\impatha{sample_000022}} & \includegraphics[width=\imwidth]{\impatha{sample_000090}} & \includegraphics[width=\imwidth]{\impathc{rdm_cartoon_sample_6}} & \includegraphics[width=\imwidth]{\impathc{rdm_cartoon_sample_7}} & \includegraphics[width=\imwidth]{\impathb{sample_000070}} & \includegraphics[width=\imwidth]{\impathb{sample_000058}} & \includegraphics[width=\imwidth]{\impathc{ic_imagenet_sample_6}} & \includegraphics[width=\imwidth]{\impathc{ic_imagenet_sample_7}} & \includegraphics[width=\imwidth]{\impathc{ic_cartoon_sample_6}} & \includegraphics[width=\imwidth]{\impathc{ic_cartoon_sample_7}} & \includegraphics[width=\imwidth]{\impathc{ic_wikiart_sample_6}} & \includegraphics[width=\imwidth]{\impathc{ic_wikiart_sample_7}} \\

\includegraphics[width=\imwidth]{\impatha{sample_000049}} & \includegraphics[width=\imwidth]{\impatha{sample_000052}} & \includegraphics[width=\imwidth]{\impathc{rdm_cartoon_sample_8}} & \includegraphics[width=\imwidth]{\impathc{rdm_cartoon_sample_9}} & \includegraphics[width=\imwidth]{\impathb{sample_000089}} & \includegraphics[width=\imwidth]{\impathb{sample_000092}} & \includegraphics[width=\imwidth]{\impathc{ic_imagenet_sample_8}} & \includegraphics[width=\imwidth]{\impathc{ic_imagenet_sample_9}} & \includegraphics[width=\imwidth]{\impathc{ic_cartoon_sample_8}} & \includegraphics[width=\imwidth]{\impathc{ic_cartoon_sample_9}} & \includegraphics[width=\imwidth]{\impathc{ic_wikiart_sample_8}} & \includegraphics[width=\imwidth]{\impathc{ic_wikiart_sample_9}} \\

\bottomrule
\end{tabular}
\caption{\label{fig:rdm_vs_icgan} Direct comparison of samples from \diffusionmodel with those of IC-GAN on i) the train-time database $\mathcal{D_{\text{\tiny train}}}$ which is the training set of ImageNet for IC-GAN and ii) on the }

\end{figure}
}

\newcommand{\ffhqrandoms}{
\renewcommand{\imwidth}{.1\textwidth}
\renewcommand{\impatha}[1]{img/ffhq_samples/##1}
\setlength{\tabcolsep}{1pt}
\begin{figure}
\centering
\begin{tabular}{c@{\hspace{0pt}}c@{\hspace{0pt}}c@{\hspace{0pt}}c@{\hspace{0pt}}c@{\hspace{0pt}}c@{\hspace{0pt}}c@{\hspace{0pt}}c@{\hspace{0pt}}c@{\hspace{0pt}}c}
\toprule
\includegraphics[width=\imwidth]{\impatha{sample_000000}}&
\includegraphics[width=\imwidth]{\impatha{sample_000001}}&
\includegraphics[width=\imwidth]{\impatha{sample_000002}}&
\includegraphics[width=\imwidth]{\impatha{sample_000003}}&
\includegraphics[width=\imwidth]{\impatha{sample_000004}}&
\includegraphics[width=\imwidth]{\impatha{sample_000005}}&
\includegraphics[width=\imwidth]{\impatha{sample_000006}}&
\includegraphics[width=\imwidth]{\impatha{sample_000007}}&
\includegraphics[width=\imwidth]{\impatha{sample_000008}}&
\includegraphics[width=\imwidth]{\impatha{sample_000009}}\\

\includegraphics[width=\imwidth]{\impatha{sample_000010}}&
\includegraphics[width=\imwidth]{\impatha{sample_000011}}&
\includegraphics[width=\imwidth]{\impatha{sample_000012}}&
\includegraphics[width=\imwidth]{\impatha{sample_000013}}&
\includegraphics[width=\imwidth]{\impatha{sample_000014}}&
\includegraphics[width=\imwidth]{\impatha{sample_000015}}&
\includegraphics[width=\imwidth]{\impatha{sample_000016}}&
\includegraphics[width=\imwidth]{\impatha{sample_000017}}&
\includegraphics[width=\imwidth]{\impatha{sample_000018}}&
\includegraphics[width=\imwidth]{\impatha{sample_000019}}\\

\includegraphics[width=\imwidth]{\impatha{sample_000020}}&
\includegraphics[width=\imwidth]{\impatha{sample_000021}}&
\includegraphics[width=\imwidth]{\impatha{sample_000022}}&
\includegraphics[width=\imwidth]{\impatha{sample_000023}}&
\includegraphics[width=\imwidth]{\impatha{sample_000024}}&
\includegraphics[width=\imwidth]{\impatha{sample_000025}}&
\includegraphics[width=\imwidth]{\impatha{sample_000026}}&
\includegraphics[width=\imwidth]{\impatha{sample_000027}}&
\includegraphics[width=\imwidth]{\impatha{sample_000028}}&
\includegraphics[width=\imwidth]{\impatha{sample_000029}}\\

\includegraphics[width=\imwidth]{\impatha{sample_000030}}&
\includegraphics[width=\imwidth]{\impatha{sample_000031}}&
\includegraphics[width=\imwidth]{\impatha{sample_000032}}&
\includegraphics[width=\imwidth]{\impatha{sample_000033}}&
\includegraphics[width=\imwidth]{\impatha{sample_000034}}&
\includegraphics[width=\imwidth]{\impatha{sample_000035}}&
\includegraphics[width=\imwidth]{\impatha{sample_000036}}&
\includegraphics[width=\imwidth]{\impatha{sample_000037}}&
\includegraphics[width=\imwidth]{\impatha{sample_000038}}&
\includegraphics[width=\imwidth]{\impatha{sample_000039}}\\

\includegraphics[width=\imwidth]{\impatha{sample_000040}}&
\includegraphics[width=\imwidth]{\impatha{sample_000041}}&
\includegraphics[width=\imwidth]{\impatha{sample_000042}}&
\includegraphics[width=\imwidth]{\impatha{sample_000043}}&
\includegraphics[width=\imwidth]{\impatha{sample_000044}}&
\includegraphics[width=\imwidth]{\impatha{sample_000045}}&
\includegraphics[width=\imwidth]{\impatha{sample_000046}}&
\includegraphics[width=\imwidth]{\impatha{sample_000047}}&
\includegraphics[width=\imwidth]{\impatha{sample_000048}}&
\includegraphics[width=\imwidth]{\impatha{sample_000049}}\\

\includegraphics[width=\imwidth]{\impatha{sample_000050}}&
\includegraphics[width=\imwidth]{\impatha{sample_000051}}&
\includegraphics[width=\imwidth]{\impatha{sample_000052}}&
\includegraphics[width=\imwidth]{\impatha{sample_000053}}&
\includegraphics[width=\imwidth]{\impatha{sample_000054}}&
\includegraphics[width=\imwidth]{\impatha{sample_000055}}&
\includegraphics[width=\imwidth]{\impatha{sample_000056}}&
\includegraphics[width=\imwidth]{\impatha{sample_000057}}&
\includegraphics[width=\imwidth]{\impatha{sample_000058}}&
\includegraphics[width=\imwidth]{\impatha{sample_000059}}\\

\includegraphics[width=\imwidth]{\impatha{sample_000060}}&
\includegraphics[width=\imwidth]{\impatha{sample_000061}}&
\includegraphics[width=\imwidth]{\impatha{sample_000062}}&
\includegraphics[width=\imwidth]{\impatha{sample_000063}}&
\includegraphics[width=\imwidth]{\impatha{sample_000064}}&
\includegraphics[width=\imwidth]{\impatha{sample_000065}}&
\includegraphics[width=\imwidth]{\impatha{sample_000066}}&
\includegraphics[width=\imwidth]{\impatha{sample_000067}}&
\includegraphics[width=\imwidth]{\impatha{sample_000068}}&
\includegraphics[width=\imwidth]{\impatha{sample_000069}}\\

\includegraphics[width=\imwidth]{\impatha{sample_000070}}&
\includegraphics[width=\imwidth]{\impatha{sample_000071}}&
\includegraphics[width=\imwidth]{\impatha{sample_000072}}&
\includegraphics[width=\imwidth]{\impatha{sample_000073}}&
\includegraphics[width=\imwidth]{\impatha{sample_000074}}&
\includegraphics[width=\imwidth]{\impatha{sample_000075}}&
\includegraphics[width=\imwidth]{\impatha{sample_000076}}&
\includegraphics[width=\imwidth]{\impatha{sample_000077}}&
\includegraphics[width=\imwidth]{\impatha{sample_000078}}&
\includegraphics[width=\imwidth]{\impatha{sample_000079}}\\

\includegraphics[width=\imwidth]{\impatha{sample_000080}}&
\includegraphics[width=\imwidth]{\impatha{sample_000081}}&
\includegraphics[width=\imwidth]{\impatha{sample_000082}}&
\includegraphics[width=\imwidth]{\impatha{sample_000083}}&
\includegraphics[width=\imwidth]{\impatha{sample_000084}}&
\includegraphics[width=\imwidth]{\impatha{sample_000085}}&
\includegraphics[width=\imwidth]{\impatha{sample_000086}}&
\includegraphics[width=\imwidth]{\impatha{sample_000087}}&
\includegraphics[width=\imwidth]{\impatha{sample_000088}}&
\includegraphics[width=\imwidth]{\impatha{sample_000089}}\\

\includegraphics[width=\imwidth]{\impatha{sample_000090}}&
\includegraphics[width=\imwidth]{\impatha{sample_000091}}&
\includegraphics[width=\imwidth]{\impatha{sample_000092}}&
\includegraphics[width=\imwidth]{\impatha{sample_000093}}&
\includegraphics[width=\imwidth]{\impatha{sample_000094}}&
\includegraphics[width=\imwidth]{\impatha{sample_000095}}&
\includegraphics[width=\imwidth]{\impatha{sample_000096}}&
\includegraphics[width=\imwidth]{\impatha{sample_000097}}&
\includegraphics[width=\imwidth]{\impatha{sample_000098}}&
\includegraphics[width=\imwidth]{\impatha{sample_000099}}\\
\bottomrule
\end{tabular}
\caption{\label{fig:ffhq_randomsamples} Random samples from our FFHQ RDM samples with 100 steps and $m=0.01$.}
\end{figure}

}

\newcommand{\generatorsuppe}{
\renewcommand{\imwidth}{.12\textwidth}
\renewcommand{\impatha}[1]{img/flow_exp/noflow/mountain/##1}
\renewcommand{\impathb}[1]{img/flow_exp/noflow/plush/##1}
\renewcommand{\impathc}[1]{img/flow_exp/flow/mountain/##1}
\renewcommand{\impathd}[1]{img/flow_exp/flow/plush/##1}
\renewcommand{\impathe}[1]{img/flow_exp/noflow/pica/##1}
\renewcommand{\impathf}[1]{img/flow_exp/flow/pica/##1}
\renewcommand{\impathg}[1]{img/flow_exp/noflow/sunset/##1}
\renewcommand{\impathh}[1]{img/flow_exp/flow/sunset/##1}
\setlength{\tabcolsep}{1pt}
\begin{figure}
\centering
\begin{tabular}{c@{\hspace{2pt}}c@{\hspace{0pt}}c c@{\hspace{0pt}}c c@{\hspace{0pt}}c c@{\hspace{0pt}}c}
&\multicolumn{2}{c}{\parbox[c]{.1\textwidth}{\centering \vspace{-.8em} \scriptsize\emph{'A photograph of a mountain.'}}} &
\multicolumn{2}{c}{\parbox[c]{.1\textwidth}{\centering \vspace{-.8em} \scriptsize\emph{'A toy zombie.'}}} &
\multicolumn{2}{c}{\parbox[c]{.1\textwidth}{\centering \vspace{-.8em} \scriptsize\emph{'A zombie in the style of Picasso.'}}} &
\multicolumn{2}{c}{\parbox[c]{.1\textwidth}{\centering \vspace{-.8em} \scriptsize\emph{'A painting of a sunset.'}}} \\

\midrule
\multirow{2}{*}{\rotatebox[origin=c]{90}{\scriptsize{no prior}}} &
\multicolumn{1}{m{\imwidth}}{\includegraphics[width=\imwidth]{\impatha{0055}}} &
\multicolumn{1}{m{\imwidth}}{\includegraphics[width=\imwidth]{\impatha{0059}}} &
\multicolumn{1}{m{\imwidth}}{\includegraphics[width=\imwidth]{\impathb{0164}}} &
\multicolumn{1}{m{\imwidth}}{\includegraphics[width=\imwidth]{\impathb{0168}}} &
\multicolumn{1}{m{\imwidth}}{\includegraphics[width=\imwidth]{\impathe{0130}}} &
\multicolumn{1}{m{\imwidth}}{\includegraphics[width=\imwidth]{\impathe{0131}}} &
\multicolumn{1}{m{\imwidth}}{\includegraphics[width=\imwidth]{\impathg{0075}}} &
\multicolumn{1}{m{\imwidth}}{\includegraphics[width=\imwidth]{\impathg{0076}}} \\ [-3pt]

&
\multicolumn{1}{m{\imwidth}}{\includegraphics[width=\imwidth]{\impatha{0060}}} &
\multicolumn{1}{m{\imwidth}}{\includegraphics[width=\imwidth]{\impatha{0063}}} &
\multicolumn{1}{m{\imwidth}}{\includegraphics[width=\imwidth]{\impathb{0171}}} &
\multicolumn{1}{m{\imwidth}}{\includegraphics[width=\imwidth]{\impathb{0174}}} &
\multicolumn{1}{m{\imwidth}}{\includegraphics[width=\imwidth]{\impathe{0132}}} &
\multicolumn{1}{m{\imwidth}}{\includegraphics[width=\imwidth]{\impathe{0133}}} &
\multicolumn{1}{m{\imwidth}}{\includegraphics[width=\imwidth]{\impathg{0077}}} &
\multicolumn{1}{m{\imwidth}}{\includegraphics[width=\imwidth]{\impathg{0078}}} \\

\midrule

\multirow{2}{*}{\rotatebox[origin=c]{90}{\scriptsize{flow prior}}} &
\multicolumn{1}{m{\imwidth}}{\includegraphics[width=\imwidth]{\impathc{0055}}} &
\multicolumn{1}{m{\imwidth}}{\includegraphics[width=\imwidth]{\impathc{0054}}} &
\multicolumn{1}{m{\imwidth}}{\includegraphics[width=\imwidth]{\impathd{0163}}} &
\multicolumn{1}{m{\imwidth}}{\includegraphics[width=\imwidth]{\impathd{0168}}} &
\multicolumn{1}{m{\imwidth}}{\includegraphics[width=\imwidth]{\impathf{0132}}} &
\multicolumn{1}{m{\imwidth}}{\includegraphics[width=\imwidth]{\impathf{0135}}} &
\multicolumn{1}{m{\imwidth}}{\includegraphics[width=\imwidth]{\impathh{0073}}} &
\multicolumn{1}{m{\imwidth}}{\includegraphics[width=\imwidth]{\impathh{0076}}} \\[-3pt]

&

\multicolumn{1}{m{\imwidth}}{\includegraphics[width=\imwidth]{\impathc{0060}}} &
\multicolumn{1}{m{\imwidth}}{\includegraphics[width=\imwidth]{\impathc{0062}}} &
\multicolumn{1}{m{\imwidth}}{\includegraphics[width=\imwidth]{\impathd{0170}}} &
\multicolumn{1}{m{\imwidth}}{\includegraphics[width=\imwidth]{\impathd{0171}}} &
\multicolumn{1}{m{\imwidth}}{\includegraphics[width=\imwidth]{\impathf{0138}}} &
\multicolumn{1}{m{\imwidth}}{\includegraphics[width=\imwidth]{\impathf{0150}}} &
\multicolumn{1}{m{\imwidth}}{\includegraphics[width=\imwidth]{\impathh{0079}}} &
\multicolumn{1}{m{\imwidth}}{\includegraphics[width=\imwidth]{\impathh{0088}}} \\

\bottomrule
\end{tabular}
\caption{\label{fig:generatorsuppe} Text-to-image generalization in CLIP latent space needs a generative prior or retrieval in order to render diverse and high-quality images. Using the CLIP text embeddings directly produces flat, non-diverse samples, whereas the normalizing flow prior clearly improves quality and diversity. See Sec.~\ref{subsec:conditional}
}
\end{figure}
}

\begin{document}

\maketitle
\vspace{-1em}
\begin{abstract}

    Novel architectures have recently improved generative image synthesis leading to excellent visual quality in various tasks. Much of this success is due to the scalability of these architectures and hence caused by a dramatic increase in model complexity and in the computational resources invested in training these models. Our work questions the underlying paradigm of compressing large training data into ever growing parametric representations. We rather present an orthogonal, semi-parametric approach. We complement comparably small diffusion or autoregressive models with a separate image database and a retrieval strategy. During training we retrieve a set of nearest neighbors from this external database for each training instance and condition the generative model on these informative samples. While the retrieval approach is providing the (local) content, the model is focusing on learning the composition of scenes based on this content. As demonstrated by our experiments, simply swapping the database for one with different contents transfers a trained model post-hoc to a novel domain. The evaluation shows competitive performance on tasks which the generative model has not been trained on, such as class-conditional synthesis, zero-shot stylization or text-to-image synthesis without requiring paired text-image data. With negligible memory and computational overhead for the external database and retrieval we can significantly reduce the parameter count of the generative model and still outperform the state-of-the-art.

\end{abstract}
\vspace{-1em}
\section{Introduction}
\vspace{-1em}
\enlargethispage{\baselineskip}
\fpp
Deep generative modeling has made tremendous leaps; especially in language modeling as well as in generative synthesis of high-fidelity images and other data types.
In particular for images, astounding results have recently been achieved~\cite{taming, adm, glide, dalle2}, and three main factors can be identified as the driving forces behind this progress:
First, the success of the transformer~\cite{transformers} has caused an architectural revolution in many vision tasks~\cite{vit}, for image synthesis especially through its combination with autoregressive modeling~\cite{taming, dalle}.
Second, since their rediscovery, diffusion models have been
applied to high-resolution image generation~\cite{sohl2015deep, score, ddpm} and, within a very short time, set new standards in generative image modeling~\cite{adm, cascadeddiff, ldm, dalle2}.
Third, these approaches \emph{scale} well~\cite{dalle,dalle2,scalinglaws, thebitterlesson}; in particular when considering the model- and batch sizes involved for high-quality \comment{diffusion }models~\cite{adm,glide,dalle,dalle2} there is evidence that this scalability is of central importance for their performance.%
However, the driving force underlying this training paradigm are models with ever growing numbers of parameters~\cite{thebitterlesson} that require huge computational resources. Besides the enormous demands in energy consumption and training time, this paradigm renders future generative modeling more and more exclusive to privileged institutions, thus hindering the democratization of research. Therefore, we here present an orthogonal approach. Inspired by recent advances in retrieval-augmented NLP~\cite{retro, DBLP:journals/corr/abs-2203-08913}, we question the prevalent approach of expensively
compressing visual concepts shared between distinct training examples into large numbers of trainable parameters and equip a comparably small generative model with a large image database. During training, our resulting \emph{semi-parametric} generative models access this database via a nearest neighbor lookup and, thus, need not learn to generate data 'from scratch'. Instead, they learn to \emph{compose} new scenes based on retrieved visual instances. This property not only increases generative performance with reduced parameter count (see Fig.~\ref{fig:fpp}), and lowers compute requirements during training.
Our proposed approach also enables the models during inference to generalize to new knowledge in form of alternative image databases without requiring further training, what can be interpreted as a form of post-hoc model modification~\cite{retro}.
We show this by replacing the retrieval database with the WikiArt~\cite{wikiart} dataset after training, thus applying the model to zero-shot stylization.

\comment{ %
However, since the huge compute resources involved in further following this paradigm by adding more training data and parameters~\cite{thebitterlesson}
render generative modeling more and more exclusive to privileged institutions,
we here present an orthogonal approach. Inspired by recent advances in retrieval-augmented NLP~\cite{retro, DBLP:journals/corr/abs-2203-08913}, we question the prevalent approach of expensively
compressing visual concepts shared between distinct training examples into trainable parameters and equip generative models with a large image database. %
During training, our resulting \emph{semi-parametric} generative models access this database via a nearest neighbor lookup and, thus, need not learn to generate data 'from scratch'. Instead, they learn to \emph{compose} new scenes based on retrieved visual instances. %
This property not only increases generative performance with reduced parameter count, \cf Fig.~\ref{fig:fpp}, and compute requirements during training; it also allows our proposed models to generalize to additionally acquired knowledge in form of alternative databases during inference, which can be interpreted as a form of post-hoc model modification~\cite{retro}.
We show this by replacing the retrieval database with the WikiArt~\cite{wikiart} dataset \emph{after} training, thus applying the model to zero-shot stylization.
}

Furthermore, our approach is formulated indepently of the underlying generative model, allowing us to present both retrieval-augmented diffusion (\diffusionmodel\!) and autoregressive (\armodel\!) models.
By searching in and conditioning on the latent space of CLIP~\cite{clip} and using scaNN~\cite{pmlr-v119-guo20h} for the NN-search, the retrieval causes negligible overheads in training/inference time (0.95 ms to retrieve 20 nearest neighbors from a database of 20M examples) and storage space (2GB per 1M examples).
We show that semi-parametric models yield high fidelity and diverse samples: \diffusionmodel surpasses recent state-of-the-art diffusion models in terms of FID and diversity while requiring less trainable parameters.
Furthermore, the shared image-text feature space of CLIP allows for various conditional applications such as text-to-image or class-conditional synthesis, despite being trained on images only (as demonstrated in Fig.~\ref{fig:clipretro}).
Finally, we present additional truncation strategies to control the synthesis process which can be combined with model specific sampling techniques such as classifier-free guidance for diffusion models~\cite{ho2021classifier} or top-$k$ sampling \cite{DBLP:journals/corr/abs-1805-04833} for autoregressive models.

\comment{
Furthermore, since our approach is formulated in a model agnostic way, we present both retrieval-augmented diffusion (\diffusionmodel) and autoregressive (\armodel) models. By searching in and conditioning on the latent space of CLIP~\cite{clip} and using scaNN~\cite{pmlr-v119-guo20h} for NN-search, we implement the nearest neighbor lookup with little overheads in training/inference time (0.95 ms to retrieve 20 nearest neighbors from a database of 20M examples) and storage space (2GB per 1M examples).
We show that semi-parametric models yield high fidelity and diverse samples: \diffusionmodel surpasses recent state-of-the-art diffusion models in terms of FID and diversity while requiring less trainable parameters.
Furthermore, the shared image-text feature space of CLIP allows for various conditional applications such as text-to-image or class-conditional synthesis, despite being trained on images only (as demonstrated in Fig.~\ref{fig:clipretro}).
Finally, we present additional truncation strategies to control the synthesis process which can be combined with model specific sampling techniques such as classifier-free guidance for diffusion models~\cite{ho2021classifier} or top-$k$ sampling \cite{DBLP:journals/corr/abs-1805-04833} for autoregressive models.
}

\vspace{-.5em}

\cliptextquerytwo

\section{Related Work}
\label{sec:related}
\vspace{-1em}
\enlargethispage{\baselineskip}

\textbf{Generative Models for Image Synthesis.}
Generating high quality novel images has long been a challenge for deep learning community due to their high dimensional nature.
Generative adversarial networks (GANs)~\cite{gan} excel at synthesizing such high resolution images with outstanding quality~\cite{biggan,stylegan2,stylegan3,stylegan-xl} while optimizing their training objective requires some sort of tricks~\cite{wassersteingan,improvedwassersteingan,mescheder2018training,numericsofgan} and their samples suffer from the lack of diversity~\cite{srivastava2017veegan,wassersteingan, unrolledgan,pacgan}.
On the contrary, likelihood-based methods have better training properties and they are easier to optimize thanks to their ability to capture the full data distribution.
While failing to achieve the image fidelity of GANs, variational autoencoders (VAEs)~\cite{vae,vae2} and flow-based methods~\cite{dinh2014nice,dinh2016density} facilitate high resolution image generation with fast sampling speed~\cite{nvae,glow}.
Autoregressive models (ARMs)~\cite{chen2020generative,pixelcnn,pixelrnn,pixelcnnplus} succeed in density estimation like the other likelihood-based methods, albeit at the expense of computational efficiency.
Starting with the seminal works of Sohl-Dickstein et al.~\cite{sohl2015deep} and Ho et al.~\cite{ddpm}, diffusion-based generative models have improved generative modeling of artificial visual systems~\cite{adm,kingma2021variational,trilemma,ho2022video,yang2022diffusion,saharia2021image}.
Their good performance, however, comes at the expense of high training costs and slow sampling.
To circumvent the drawbacks of ARMs and diffusion models, several two-stage models are proposed to scale them to higher resolutions by training them on the compressed image features~\cite{vqvae,vqvae2,taming,yu2021vectorquantized,ldm,d2c,imagebart}.
However, they still require large models and significant compute resources, especially for unconditional image generation~\cite{adm} on complex datasets like ImageNet~\cite{imagenet} or complex conditional tasks such as text-to-image generation~\cite{glide,dalle,gu2021vector,ldm}.
To address these issues, given limited compute resources, we propose to trade trainable parameters for an external memory which empowers smaller models to achieve high fidelity image generation.%

\textbf{Retrieval-Augmented Generative Models.} Using external memory to augment traditional models has recently drawn attention in natural language processing (NLP)~\cite{khandelwal2019generalization,khandelwal2020nearest,meng2021gnn,guu2020retrieval}.
For example, RETRO~\cite{retro} proposes a retrieval-enhanced transformer for language modeling which performs on par with state-of-the-art models~\cite{gpt3} using significantly less parameters and compute resources.
These retrieval-augmented models with external memory turn purely parametric deep learning models into semi-parametric ones.
Early attempts~\cite{long2022retrieval,siddiqui2021retrievalfuse,tseng2020retrievegan,xu2021texture} in retrieval-augmented visual models do not use an external memory and exploit the training data itself for retrieval.
In image synthesis, IC-GAN~\cite{ic-gan} utilizes the neighborhood of training images to train a GAN and generates samples by conditioning on single instances from the training data.
However, using training data itself for retrieval potentially limits the generalization capacity, and thus, we favor an external memory in this work.

\section{Image Synthesis with Retrieval-Augmented Generative Models}
\label{sec:mehthod}
\modelfigure
\vspace{-1em}
\enlargethispage{\baselineskip}
Our work considers data points as an explicit \emph{part of the model}. In contrast to common neural generative approaches for  image synthesis~\cite{biggan,stylegan3,stylegan-xl,vqvae2,taming,chen2020generative,chai2022any}, this approach is not only parameterized by the learnable weights of a neural network, but also
a (fixed) set of data representations and a non-learnable \emph{retrieval} %
function, which, given a query from the training data, retrieves suitable data representations from the external dataset. Following prior work in natural language modeling~\cite{retro},
we implement this retrieval pipeline as a nearest neighbor lookup. %

Sec.~\ref{subsec:methodintro} and Sec.~\ref{sec:model_training} formalize this approach for training retrieval-augmented diffusion and autoregressive models for image synthesis, while Sec.~\ref{subsec:model_inference} %
introduces sampling mechanisms that become available once such a model is trained. Fig.~\ref{fig:model} provides an overview over our approach.
\vspace{-.5em}
\subsection{Retrieval-Enhanced Generative Models of Images}
\label{subsec:methodintro}
\vspace{-.8em}
Unlike common, fully parametric neural generative approaches for images, we define a \emph{semi-parametric} generative image model $p_{\theta,\mathcal D, \sampling}(x)$ by introducing trainable parameters $\theta $ \emph{and} non-trainable model components $\mathcal{D}, \sampling$,  where $\mathcal{D} = \{y_i\}_{i=1}^{N}$ is a \emph{fixed} database of images $y_i \in \mathbb{R}^{H_{\mathcal{D}} \times W_{\mathcal{D}} \times 3}$ that is disjoint from our train data $\mathcal X$. %
Further, $\sampling$ denotes a (non-trainable) sampling strategy to obtain a subset of $\mathcal{D}$ based on a query
$x$, i.e. $\sampling\colon x, \mathcal{D} \mapsto \mathcal{M}_{\mathcal{D}}^{(k)}$, where $\mathcal{M}_{\mathcal{D}}^{(k)} \subseteq \mathcal{D}$ and $\vert \mathcal{M}_{\mathcal{D}}^{(k)} \vert = k$ .
Thus, only $\theta$ is actually learned during training.

Importantly, $\sampling(x, \mathcal{D})$ has to be chosen such that it provides the model with beneficial visual representations from $\mathcal{D}$ for modeling $x$ and the entire capacity of $\theta$ can be leveraged to \emph{compose} consistent scenes based on these patterns.
For instance, considering query images $x \in \mathbb{R}^{H_x \times W_x \times 3}$, a valid strategy $\sampling(x, \mathcal{D})$ is a function that for each $x$ returns the set of its $k$ nearest neighbors, measured by a given distance function $d(x,\cdot)$.

Next, we propose to provide this retrieved information to the model via \emph{conditioning}, i.e. we %
specify a general semi-parametric generative model as
\begin{equation}
    p_{\theta, \mathcal{D}, \sampling}(x)
    = p_{\theta}\big(x \mid \sampling(x, \mathcal{D})\big)
    = p_{\theta}(x \mid \mathcal{M}_{\mathcal{D}}^{(k)})
\label{eq:modelbase}
\end{equation}
In principle, one could directly use image samples $y \in \mathcal{M}_{\mathcal{D}}^{(k)}$ to learn $\theta$.
However, since images contain many ambiguities and their high dimensionality involves considerable computational and storage cost\footnote{Note that $\mathcal{D}$ is essentially a part of the model weights} we use a \emph{fixed}, pre-trained image encoder $\phi$ to project all examples from $\mathcal{M}_{\mathcal{D}}^{(k)}$ onto a low-dimensional manifold. Hence, Eq.~\eqref{eq:modelbase} reads
 \begin{equation}
 p_{\theta, \mathcal{D}, \sampling}(x)
 = p_{\theta}(x \mid \{\,\phi(y) \mid y \in \sampling(x, \mathcal{D})\,\}).
 \label{eq:model}
 \end{equation}

where $p_\theta(x \vert \cdot)$ is a conditional generative model with trainable parameters $\theta$ which we refer to as \emph{decoding head}. With this, the above procedure can be applied to any type of generative decoding head and is not dependent on its concrete training procedure.

\vspace{-.8em}
\subsection{Instances of Semi-Parametric Generative Image Models}
\label{sec:model_training}
\vspace{-.8em}
\enlargethispage{\baselineskip}
During training we are given a train dataset $\mathcal{X} = \{x_i\}_{i=1}^{M}$ of images whose distribution $p(x)$ we want to approximate with $p_{\theta,\mathcal D, \sampling}(x)$. Our train-time
sampling strategy $\sampling$ uses a query example $x \sim p(x)$ to retrieve its $k$ nearest neighbors $y \in \mathcal{D}$ by implementing $d(x,y)$ as the cosine similarity in the image feature space of CLIP~\cite{clip}.
Given a sufficiently large database $\mathcal{D}$, this strategy ensures that the set of neighbors $\sampling(x, \mathcal{D})$ shares sufficient information with $x$ and, thus, provides useful visual information for the generative task.
We choose CLIP to implement $\sampling$, because it embeds images in a low dimensional space ($\dim = 512$) and maps semantically similar samples to the same neighborhood, yielding an efficient search space. Fig.~\ref{fig:example_nns} visualizes examples of nearest neighbors retrieved via a ViT-B/32 vision transformer~\cite{vit} backbone.
\nnsexamples

Note that this approach can, in principle, turn any generative model into a semi-parametric model in the sense of Eq.~\eqref{eq:model}. In this work we focus on models where the decoding head is either implemented as a diffusion or an autoregressive model, motivated by the success of these models in image synthesis~\cite{ddpm,adm,ldm,glide,dalle,taming}.

To obtain the image representations via $\phi$, different encoding models are conceivable in principle.
Again, the latent space of CLIP offers some advantages since it is (i) very compact, which (ii) also reduces memory requirements. Moreover, the contrastive pretraining objective (iii) provides a shared space of image and text representations, which is beneficial for text-image synthesis, as we show in Sec.~\ref{subsec:conditional}.
Unless otherwise specified, $\phi \equiv \phi_{\text{CLIP}}$ is set in the following. We investigate alternative parameterizations of $\phi$ in Sec.~\ref{suppsubsec:sp_rep}.

Note that with this choice, the additional database $\mathcal{D}$ can also be interpreted as a fixed \emph{embedding layer}\footnote{For a database of 1M images and using 32-bit precision, this equals approximately 2.048 GB} of dimensionality $\vert \mathcal{D} \vert \times 512$ from which the nearest neighbors are retrieved.

\vspace{-.8em}
\subsubsection{Retrieval-Augmented Diffusion Models}
\label{subsubsec:diffusionspecific}
\vspace{-.5em}
\enlargethispage{\baselineskip}
In order to reduce computational complexity and memory requirements during training, we follow~\cite{ldm} and build on latent diffusion models (LDMs) which learn the data distribution in the latent space $z=E(x)$ of a pretrained autoencoder.
We dub this retrieval-augmented latent diffusion model \diffusionmodel and train it with the usual reweighted likelihood objective~\cite{ddpm}, yielding the objective~\cite{sohl2015deep,ddpm}
\begin{equation}
\label{eq:loss}
\min_{\theta} \mathcal{L} = \mathbb{E}_{p(x), z \sim E(x), \epsilon \sim \mathcal{N}(0,1), t}\Big[ \Vert \epsilon - \epsilon_{\theta}\big(z_t,\, t,\, \{\phi_{\text{CLIP}}(y) \mid y \in \sampling(x, \mathcal{D})\}\big) \Vert_2^{2} \Big] \, ,
\end{equation}
where the expectation is approximated by the empirical mean over training examples.
In the above equation, $\epsilon_\theta$ denotes the UNet-based~\cite{DBLP:conf/miccai/RonnebergerFB15} denoising autoencoder as used in~\cite{adm,ldm} and $t \sim \text{Uniform}\{1,\ldots,T\}$ denotes the time step~\cite{sohl2015deep,ddpm}. To feed the set of  nearest neighbor encodings $\phi_{\text{CLIP}}(y)$ into $\epsilon_\theta$, we use the cross-attention conditioning mechanism proposed in~\cite{ldm}.
\vspace{-.8em}
\subsubsection{Retrieval-Augmented Autoregressive Models}
\label{sec:model_training_ar}
\vspace{-.5em}
Our approach is applicable to several types of likelihood-based methods. We show this by augmenting diffusion models (Sec.~\ref{subsubsec:diffusionspecific}) as well as autoregressive models with the retrieved representations. To implement the latter, we follow \cite{taming} and train autoregressive transformer models to model the distribution of the discrete image tokens $z_q=E(x)$ of a VQGAN \cite{taming, vqvae}. Specifically, as for \diffusionmodel, we train retrieval-augmented autoregressive models \emph{(RARMs)} conditioned on the CLIP embeddings $\phi_{\text{CLIP}}(y)$ of the neighbors $y$, so that the objective reads
\begin{equation}
\min_{\theta} \mathcal{L} = - \mathbb{E}_{p(x), z_q \sim E(x)} \Big[ \sum_i \log p\big(z_q^{(i)} \mid z_q^{(<i)},\, \{\phi_{\text{CLIP}}(y) \mid y \in \sampling(x, \mathcal{D})\}\big) \Big] \, ,
\label{eq:rarm}
\end{equation}
where we choose a row-major ordering for the autoregressive factorization of the latent $z_q$.
We condition the model on the set of neighbor embeddings $\phi_{\text{CLIP}}(\sampling(x, \mathcal{D}))$ via cross-attention~\cite{transformers}.
\vspace{-0.5em}
\subsection{Inference for Retrieval-Augmented Generative Models}
\label{subsec:model_inference}
\vspace{-.5em}
\paragraph{Conditional Synthesis without Conditional Training}
Being able to change the (non-learned) $\mathcal{D}$ and $\sampling$ at test time offers additional flexibility compared to standard generative approaches:
Depending on the application, it is possible to extent/restrict $\mathcal{D}$ for particular exemplars; or to skip the retrieval via $\sampling$ altogether and provide a set of representations $\{\phi_{\text{CLIP}}(y_i)\}_{i=1}^k$ directly.
This allows us to use additional conditional information such as a text prompt or a class label, which has not been available during training, to achieve more fine-grained control during synthesis.

For \textbf{text-to-image generation}, for example, our model can be conditioned in several ways:
Given a text prompt $c_{\text{text}}$ and using the text-to-image retrieval ability of CLIP,
we can retrieve $k$ neighbors from $\mathcal{D}$ and use these as an implicit text-based conditioning. However, since we condition on CLIP representations $\phi_{\text{CLIP}}$, we can also condition directly on the \emph{text} embeddings obtained via CLIP's language backbone (since CLIP's text-image embedding space is shared). Accordingly, it is also possible to combine these approaches and use text and image representations simultaneously. We show and compare the results of using these sampling techniques in Fig.~\ref{fig:clipretro}.

Given a class label $c$, we define a text such as \emph{'An image of a $t(c)$.'} based on its textual description $t(c)$ or apply the embedding strategy for text prompts and sample a pool $\xi_l(c) \, , \, k \leq l$ for each class. By randomly selecting $k$ adjacent examples from this pool for a given query $c$, we obtain an inference-time class-conditional model and analyze these post-hoc conditioning methods in Sec.~\ref{subsec:conditional}.

For \textbf{unconditional generative modeling}, we randomly sample a pseudo-query $\tilde{x} \in \mathcal{D}$
to obtain the set $\sampling^{\text{test}}(\tilde{x}, \mathcal{D})$ of its $k$ nearest neighbors. Given this set,
 Eq.~\eqref{eq:model} can be used to draw samples, since $p_{\theta}(x \vert \cdot)$ itself is a generative model. However, when generating all samples from $p_{\theta,\mathcal D, \sampling}(x)$ only from one particular set $\sampling^{\text{test}}(\tilde{x})$, we expect $p_{\theta,\mathcal D, \sampling}(x)$ to be unimodal and sharply peaked around $\tilde{x}$. %
When intending to model a complex multimodal distribution $p(x)$ of natural images, this choice would obviously lead to weak results. Therefore, we
construct
a proposal distribution based on $\mathcal{D}$
where
\begin{equation}
    p_{\mathcal{D}}(\tilde{x}) = \frac{\lvert \{x \in \mathcal X \mid \tilde{x} \in \sampling(x, \mathcal{D})\}\rvert}{k \cdot \lvert \mathcal X \rvert}\;,
    \quad \text{for} \; \tilde{x} \in \mathcal{D}\;.
\end{equation}

This definition counts the instances in the database $\mathcal{D}$ which are useful for modeling the training dataset $\mathcal{X}$.
Note that $p_{\mathcal{D}}(\tilde{x})$ only depends on $\mathcal{X}$ and $\mathcal{D}$, what allows us to precompute it. Given $p_{\mathcal{D}}(\tilde{x})$, we can obtain a set
\begin{equation}
\label{eq:uncond_sampling}
\mathcal{P} = \Big\{x \sim p_{\theta}(x \mid \{\,\phi(y) \mid y \in \sampling(\tilde{x}, \mathcal{D})\,\}) \ \Big|\ \tilde{x} \sim p_{\mathcal{D}}(\tilde{x}) \Big\} \,
\end{equation}
of samples from the our model. We can thus draw from the unconditional modeled density $p_{\theta,\mathcal D, \sampling}(x)$ by drawing $x \sim \text{Uniform}(\mathcal{P})$.

By choosing only a fraction $m \in (0, 1]$ of most likely examples $\tilde{x} \sim p_{\mathcal{D}}(\tilde{x})$, we can artificially truncate this distribution and trade sample quality for diversity. See Sec.~\ref{suppsubsec:topm_details}.
for a detailed description of this mechanism which we call \emph{top-m sampling} and Sec.~\ref{subsec:sampling_hacks} for an empirical demonstration.
\sampleswithnns
\vspace{-.5em}
\section{Experiments}
\vspace{-.5em}
\vspace{-0.5em}
This section presents experiments for both retrieval-augmented diffusion and autoregressive models.
To obtain nearest neighbors we apply the ScaNN search algorithm~\cite{pmlr-v119-guo20h} in the feature space of a pretrained CLIP-ViT-B/32~\cite{clip}. Using this setting, retrieving 20 nearest neighbors from the database described above takes $\sim 0.95$ ms. For more details on our retrieval implementation, see Sec.~\ref{suppsubsec:retrieval}.
For quantitative performance measures we use FID~\cite{FID}, CLIP-FID~\cite{clip-fid}, Inception Score (IS)~\cite{Salimans2016ImprovedTF} and Precision-Recall~\cite{DBLP:journals/corr/abs-1904-06991}, and, for the diffusion models, generate samples with the DDIM sampler~\cite{ddim} with 100 steps and $\eta = 1$. For hyperparameters, implementation and evaluation details \cf Sec.~\ref{suppsec:implementation_details}.

\vspace{-.5em}
\subsection{Semi-Parametric Image Generation}
\label{subsec:unconditional}
\vspace{-.5em}
Drawing pseudo-queries from the proposal distribution proposed in Sec.~\ref{subsec:model_inference} and Eq.~\eqref{eq:uncond_sampling} enables semi-parametric unconditional image generation. However, before the actual application, we compare different choices of the database $\mathcal{D}_{\text{\tiny train}}$ used during training and determine an appropriate choice for the value $k$ of the retrieved neighbors during training.

\traindatabaseablation
\textbf{Finding a train-time database $\mathcal{D}_{\text{\tiny train}}$.}
Key to a successful application of semi-parametric models is choosing an appropriate train database $\mathcal{D}_{\text{\tiny train}}$, as it has to provide the generative backbone $p_{\theta}$ with useful information.
We hypothesize that a large database with diverse visual instances is most useful for the model, since the probability of finding nearby neighbors in $\mathcal{D}_{\text{\tiny train}}$ for \emph{every} train example is highest for this choice. To verify this claim, we compare the visual quality and sample diversity of three \diffusionmodels trained on the dogs-subset of ImageNet~\cite{imagenet} with i) WikiArt~\cite{wikiart} (\wadiffusion\!), ii) MS-COCO~\cite{DBLP:conf/cvpr/CaesarUF18} (\cocodiffusion\!) and iii) 20M examples obtained by cropping images (see App.~\ref{suppsubsec:retrieval}) from OpenImages~\cite{DBLP:journals/corr/abs-1811-00982} (\oidiffusion\!) as train database $\mathcal{D}_{\text{\tiny train}}$ with that of an \emph{LDM} baseline with 1.3$\times$ more parameters. Fig~\ref{fig:dataset_ablation_metrics} shows that i) a database $\mathcal{D}_{\text{\tiny train}}$, whose examples are from a different domain than those of the train set $\mathcal{X}$ leads to degraded sample quality, whereas ii) a small database from the same domain as $\mathcal{X}$ improves performance compared to the \emph{LDM} baseline. Finally, iii) increasing the size of $\mathcal{D}_{\text{\tiny train}}$ further boosts performance in quality and diversity metrics and leads to significant improvements of \diffusionmodels compared to \emph{LDMs}.

\databasegeneralization
For the above experiment we used $\mathcal{D}_{\text{\tiny train}} \cap \mathcal{X} = \emptyset$. This is in contrast to prior work~\cite{ic-gan} which conditions a generative model on the train dataset itself, i.e., $\mathcal{D}_{\text{\tiny train}} = \mathcal{X}$. Our choice is motivated by the aim to obtain a model as general as possible which can be used for more than one task during inference, as introduced in Sec.~\ref{subsec:model_inference}. To show the benefits of using $\mathcal{D}_{\text{\tiny train}} \cap \mathcal{X} = \emptyset$ we use ImageNet~\cite{imagenet} as train set $\mathcal{X}$ and compare \oidiffusion with an \diffusionmodel conditioned on $\mathcal{X}$ itself (\indiffusion\!). We evaluate their performance on the ImageNet train- and validation-sets in Tab.~\ref{tab:database_generalization}, which shows \oidiffusion to closely reach the performance of \indiffusion in CLIP-FID~\cite{clip-fid} and achieve more diverse results. When interchanging the test-time database between the two models, i.e., conditioning \oidiffusion on examples from ImageNet (\oiindiffusion\!) and vice versa (\inoidiffusion\!) we observe strong performance degradation of the latter model, whereas the former improves in most metrics and outperforms \indiffusion in CLIP-FID, thus showing the enhanced generalization capabilities when choosing $\mathcal{D}_{\text{\tiny train}} \cap \mathcal{X} = \emptyset$. To provide further evidence of this property we additionally evaluate the models on zero-shot text-conditional on the COCO dataset~\cite{DBLP:conf/cvpr/CaesarUF18} in Tab.~\ref{tab:database_generalization}. Again, we observe better image quality (FID) as well as image-text alignment (CLIP-score) of \oidiffusion which furthermore outperforms LAFITE~\cite{zhou2021lafite} in FID, despite being trained on only a third of the train examples.

\knnwrapped
\textbf{How many neighbors to retrieve during training?}
As the number $k_{\text{train}}$ of retrieved nearest neighbors during training has a
strong influence on the properties of the resulting
model after training, we
first identify hyperparameters obtain a model with optimal synthesis properties.
Hence, we parameterize $p_\theta$ with a diffusion model and train five models for different $k_{\text{train}} \in \{1,2,4,8,16\}$
on ImageNet~\cite{imagenet}. All models use identical generative backbones and computational resources (details in Sec.~\ref{suppsec:implementation_details_rdm}).
Fig.~\ref{fig:knn_wrapped} shows resulting performance metrics assessed on 1000 samples. For FID and IS we do not observe significant trends. Considering precision and recall, however, we see that increasing $k_{\text{train}}$ trades consistency for diversity. Large $k_{\text{train}}$ causes recall, i.e. sample diversity, to deteriorate again.

We attribute this to a regularizing influence of non-redundant, additional information beyond the single nearest neighbor, which is fed to the respective model during training, when $k_{\text{train}} > 1$. For $k_{\text{train}} \in \{2,4,8\}$ this additional information is beneficial and the corresponding models appropriately mediate between quality and diversity. Thus, we use $k=4$ for our main \diffusionmodel.
Furthermore, the numbers of neighbors has a significant effect on the generalization capabilities of our model for \emph{conditional} synthesis, \eg text-to-image synthesis as in Fig.~\ref{fig:clipretro}. We provide an in-depth evaluation of this effect in Sec.~\ref{subsec:conditional} and conduct a similar study for \armodel in Sec.~\ref{suppsubsec:knn_rarm}.
\enlargethispage{\baselineskip}

\textbf{Qualitative results.}
Fig.~\ref{fig:samples_with_nns} shows samples of \diffusionmodel/\armodel trained on ImageNet as well as \diffusionmodel samples on FFHQ~\cite{stylegan1} for different sets $\mathcal{M}_{\mathcal{D}}^{(k)}(\tilde{x})$ of retrieved neighbors given a pseudo-query $\tilde{x} \sim p_{\mathcal{D}}(\tilde{x})$. We also plot the nearest neighbors from the train set to show that this set is disjoint from the database $\mathcal{D}$ and that our model renders new, unseen samples. %

\textbf{Quantitative results.}
Tab.~\ref{tab:uin_metrics} compares our model with the recent state-of-the-art diffusion model ADM~\cite{adm} and the semi-parametric GAN-based model IC-GAN~\cite{ic-gan} (which requires access to the \emph{training set} examples during inference) in unconditional image synthesis on ImageNet~\cite{imagenet} $256 \times 256$. %

\uinmetrics
To boost performance, we use the sampling strategies proposed in Sec.~\ref{subsec:model_inference} (which is also further detailed in Sec.~\ref{suppsubsec:topm_details}). %
With classifier-free guidance (c.f.g.), our model attains better scores than IC-GAN and ADM while being on par with ADM-G~\cite{adm}. The latter requires an additional classifier and the labels of training instances during inference.
Without any additional information about training data, e.g., image labels, \diffusionmodel achieves the best overall performance.%

\ffhqmetrics
For $m=0.1$, our retrieval-augmented diffusion model surpasses unconditional ADM for FID, IS, precision and, without guidance, for recall.
For $s=1.75$, we observe bisected FID scores compared to our unguided model and even reach the guided model ADM-G, which, unlike \diffusionmodel, requires a classifier that is pre-trained on noisy data representations.
The optimal parameters for FID are $m=0.05,\, s=1.5$, as in the bottom row of Tab.~\ref{tab:uin_metrics}. Using these parameters for \indiffusion results in a model which even achieves similar FID scores than state of the class-conditional models on ImageNet~\cite{ldm,adm,stylegan-xl} without requiring any labels during training or inference. %
Overall, this shows the strong performance of \diffusionmodel and the flexibility of top-m sampling and c.f.g., which we further analyze in Sec.~\ref{subsec:sampling_hacks}. Moreover we train an exact replicate of our ImageNet \oidiffusion on the FFHQ~\cite{stylegan1} and summarize the results in Tab.~\ref{tab:ffhq_metrics}. Since FID~\cite{FID} has been shown to be ``insensitive to the facial region''~\cite{clip-fid} we again use CLIP-based metrics.
Even for this simple dataset, our retrieval-based strategy proves beneficial, outperforming strong GAN and diffusion baselines, albeit at the cost of lower diversity (recall).

\vspace{-.5em}
\subsection{Conditional Synthesis without Conditional Training}
\label{subsec:conditional}
\vspace{-.5em}
\enlargethispage{\baselineskip}
\wrappedgeneralization
\paragraph{Text-to-Image Synthesis} In Fig.~\ref{fig:clipretro}, we show the zero-shot text-to-image synthesis capabilities of our ImageNet model for user defined text prompts. When building the set $\mathcal{M}_{\mathcal{D}}^{(k)}(c_{\text{text}})$ by directly using \emph{i)} the CLIP encodings $\phi_{\text{CLIP}}(c_{\text{text}})$ of the actual textual description itself (top row), we interestingly see that our model generalizes to generating fictional descriptions and transfers attributions across object classes. However, when using \emph{ii)} $\phi_{\text{CLIP}}(c_{\text{text}})$ together with its $k-1$ nearest neighbors from the database $\mathcal{D}$ as done in~\cite{knn-diff}, the model does not generalize to these difficult conditional inputs (mid row). When \emph{iii)} only using the $k$ CLIP image representations of the nearest neighbors, the results are even worse (bottom row).
We evaluate the text-to-image capabilities of \diffusionmodels on 30000 examples from the COCO validation set and compare with LAFITE~\cite{zhou2021lafite}. The latter is also based on CLIP space, but unlike our method, the image features are translated to text features by utilizing a supervised model in order to address the mismatch between CLIP text and image features. Tab.~\ref{tab:database_generalization} summarizes the results and shows that our \oidiffusion obtains better image quality as measured by the FID score.

Similar to Sec.~\ref{subsec:unconditional} we investigate the influence of $k_{\text{train}}$ on the text-to-image generalization capability of \diffusionmodel\!. To this end we evaluate the zero-shot transferability of the ImageNet models presented in the last
section to text-conditional image generation and, using strategy i) from the last paragraph, evaluate their performance on 2000 captions from the validation set of COCO~\cite{DBLP:conf/cvpr/CaesarUF18}. Fig.~\ref{fig:generalization_quant} compares the resulting FID and CLIP scores on COCO for the different choices of $k_{\text{train}}$. As a reference for the train performance, we furthermore plot the ImageNet FID. Similar to Fig.~\ref{fig:knn_wrapped} we find that small $k_{\text{train}}$ lead to weak generalization properties,
since the corresponding models cannot handle misalignments between the text representation received during inference and image representations it is trained on.
Increasing $k_{\text{train}}$ results in sets $\mathcal{M}_{\mathcal{D}}^{(k)}(x)$  which cover a larger feature space volume, what regularizes the corresponding models to be more robust against such misalignments.
Consequently, the generalization abilities increase with $k_{\text{train}}$ and reach an optimum at $k_{\text{train}} = 8$. Further increasing $k_{\text{train}}$ results in decreased information  provided via the retrieved neighbors (\cf Fig.~\ref{fig:example_nns}) and causes deteriorating generalization capabilities.

\generator
We note the similarity of this approach to \cite{dalle2}, which, by directly conditioning on the CLIP image representations of the data, essentially learns to invert the abstract image embedding.
In our framework, this corresponds to $\sampling (x) = \phi_{\text{CLIP}}(x)$ (i.e., no external database is provided). In order to fix the misalignment between text embeddings and image embeddings, \cite{dalle2} learns a conditional diffusion model for the generative mapping between these representations, requiring paired data. We argue that our retrieval-augmented approach provides an orthogonal approach to this task \emph{without} requiring paired data.
To demonstrate this, we train an ``inversion model'' as described above, i.e., use  $\sampling (x) = \phi_{\text{CLIP}}(x)$ with the same number of trainable parameters and computational budget as for the study in Fig.~\ref{fig:generalization_quant}. When directly using text embeddings for inference, the model renders samples which generally resemble the prompt, but the visual quality is low (CLIP score $0.26 \pm 0.05$, FID $\sim 87$). Modeling the prior with a conditional normalizing flow \cite{DBLP:conf/iclr/DinhSB17, DBLP:conf/nips/RombachEO20} improves the visual quality and achieves similar results in terms of text-consistency (CLIP score $0.26 \pm 0.3$, FID $\sim 45$), albeit requiring paired data. See Fig.~\ref{fig:generator} for a qualitative visualization and Appendix \ref{suppsec:implementation_details_rdm} for implementation and training details.
\ccsamples
\paragraph{Class-Conditional Synthesis} Similarly we can apply our model to zero-shot class-conditional image synthesis as proposed in Sec.~\ref{subsec:model_inference}. Fig.~\ref{fig:zero_shot_cc} shows samples from our model for classes from ImageNet. More samples for all experiments can be found in Sec.~\ref{suppsec:additional_samples}.

\vspace{-0.5em}
\subsection{Zero-Shot Text-Guided Stylization by Exchanging the Database}
\label{subsec:stylization}
\vspace{-.5em}
\stylizer
In our semi-parametric model, the retrieval database $\mathcal{D}$ is an explicit part of the synthesis model. This allows novel applications, such as replacing this database after training to modify the model and thus its output.
In this section we replace $\mathcal{D{\text{\tiny train}}}$ of the ImageNet-\diffusionmodel built from OpenImages with an alternate database $\mathcal{D_{\text{\tiny style}}}$, which contains all 138k images of the WikiArt dataset~\cite{wikiart}. As in Sec.~\ref{subsec:conditional} we retrieve neighbors from $\mathcal{D_{\text{\tiny style}}}$ via a text prompt and use the text-retrieval strategy \emph{iii)}. %
Results are shown in Fig.~\ref{fig:stylize} (top row). Our model, though only trained on ImageNet, generalizes to this new database and is capable of generating artwork-like images %
which depict the content defined by the text prompts.
To further emphasize the effects of this post-hoc exchange of $\mathcal{D}$, we show samples obtained with the same procedure but using $\mathcal{D{\text{\tiny train}}}$ (bottom row).
\vspace{-.5em}
\subsection{Increasing Dataset Complexity}
\label{subsec:dataset_complexity}
\vspace{-.2em}
\enlargethispage{\baselineskip}
To investigate their versatility for complex generative tasks, we compare semi-parametric models to their fully-parametric counterparts when systematically increasing the complexity of the training data $p(x)$. For both \diffusionmodel and \armodel\!, we train three identical models and corresponding fully parametric baselines %
(for details \cf Sec.~\ref{suppsubsec:training}) on the dogs-, mammals- and animals-subsets of ImageNet~\cite{imagenet}, \cf Tab.~\ref{tab:in_subsets}, until convergence.
Fig.~\ref{fig:dataset_complexity} visualizes the results. Even for lower-complexity datasets such as \emph{IN-Dogs}, our semi-parametric models improve over the baselines except for recall, where \armodel performance slightly worse than a standard AR model. For more complex datasets, the performance gains become more significant. Interestingly, the recall scores of our models \emph{improve} with increasing complexity, while those of the baselines strongly degrade. We attribute this to the explicit access of semi-parametric models to nearby visual instances for \emph{all} classes including underrepresented ones via the $p_{\mathcal{D}}(\tilde{x})$, \cf Eq.~\eqref{eq:uncond_sampling}, whereas a standard generative model might focus only on the modes containing the most often occurring classes (dogs in the case of ImageNet).
\datasetcomplexity
\vspace{-.5em}
\subsection{Quality-Diversity Trade-Offs}
\label{subsec:sampling_hacks}
\textbf{Top-m sampling.}
In this section, we evaluate the effects of the \emph{top-m sampling} strategy introduced in Sec.~\ref{subsec:model_inference}.
We train a \diffusionmodel on the ImageNet~\cite{imagenet} dataset
and assess the usual generative performance metrics based on 50k generated
samples and the entire training set~\cite{biggan}. Results are shown in Fig.~\ref{fig:topm_trunc_wrapped}.
For precision and recall scores, we observe a truncation behavior similar to other inference-time sampling techniques~\cite{biggan,adm, ho2021classifier,DBLP:journals/corr/abs-1805-04833}: For small values of $m$, we obtain coherent samples, which all come from a single or a small number of modes, as indicated by large precision scores. Increasing $m$, on the other hand, boosts diversity at the expense of consistency. For FID and IS, we find a sweet spot for $m=0.01$, which yields optima for both of these metrics. Visual examples for different values of $m$ are shown in
the Fig.~\ref{fig:topm_visual}.
Sec.~\ref{suppsubsec:rarm_topm} also contains similar experiments for \armodel. %
\qualitydiversitymerged

\textbf{Classifier-free guidance.}
Since \diffusionmodel is a conditional diffusion model (conditioned on the neighbor encodings $\phi(y)$), we can
apply classifier-free diffusion guidance~\cite{ho2021classifier} also for unconditional modeling.
Interestingly, we find that we can apply this technique without adding an
additional $\emptyset$-label to account for a purely unconditional setting while training $\epsilon_\theta$, as originally proposed in~\cite{ho2021classifier} and instead use a vector of zeros to generate an unconditional prediction with $\epsilon_\theta$. %
Additionally, this technique can be combined with \emph{top-m sampling} to obtain further control during sampling. In Fig.~\ref{fig:retro_guiding_wrapped} we show the effects of this combination for the ImageNet-model as described in the previous paragraph, with $m \in \{0.01, 0.1\}$ and classifier scale $s \in \{1.0, 1.25, 1.5, 1.75, 2.0, 3.0\}$, from left to right for each line. Moreover we qualitatively show the effects of guidance in Fig.~\ref{fig:retro_guiding_samples},
demonstrating the versatility of these sampling strategies during inference.

\vspace{-1em}
\section{Conclusion}
\vspace{-1em}
\enlargethispage{\baselineskip}
This paper questions the prevalent paradigm of current generative image synthesis: rather than compressing large training data in ever-growing generative models, we have proposed to efficiently store an image database and condition a comparably small generative model directly on meaningful samples from the database. To identify informative samples for the synthesis tasks at hand we follow an efficient retrieval-based approach. In the experiments our approach has outperformed the state of the art on various synthesis tasks despite demanding significantly less memory and compute. Moreover, it allows (i) conditional synthesis for tasks for which it has not been explicitly trained, and (ii) post-hoc transfer of a model to new domains by simply replacing the retrieval database.
Combined with CLIP's joint feature space, our model achieves strong results on text-image synthesis, despite being trained only on images. In particular, our retrieval-based approach eliminates the need to train an explicit generative prior model in the latent CLIP space by directly covering the neighborhood of a given data point.
While we assume that our approach still benefits from scaling, it shows a path to more efficiently trained generative models of images.
\vspace{-1em}
\section*{Acknowledgements}
\vspace{-1em}
This work has been funded by the Deutsche Forschungsgemeinschaft (DFG, German Research Foundation) within project 421703927 and the German Federal Ministry for Economic Affairs and Energy within the project KI-Absicherung - Safe AI for automated driving.

\clearpage
\newpage
{\small
\bibliographystyle{plainnat}
\bibliography{ms}
}

\clearpage
\section*{Checklist}

\begin{enumerate}
  \item For all authors...
  \begin{enumerate}
    \item Do the main claims made in the abstract and introduction accurately reflect the paper's contributions and scope?
      \answerYes{}
    \item Did you describe the limitations of your work?
      \answerYes{See the supplemental material.}
    \item Did you discuss any potential negative societal impacts of your work?
      \answerYes{See the supplemental material.}
    \item Have you read the ethics review guidelines and ensured that your paper conforms to them?
      \answerYes{}
  \end{enumerate}
  \item If you are including theoretical results...
  \begin{enumerate}
    \item Did you state the full set of assumptions of all theoretical results?
      \answerNA{}
    \item Did you include complete proofs of all theoretical results?
      \answerNA{}
  \end{enumerate}
  \item If you ran experiments...
  \begin{enumerate}
    \item Did you include the code, data, and instructions needed to reproduce the main experimental results (either in the supplemental material or as a URL)?
      \answerYes{The code will be released, the data is publicly available and the additional instructions are provided in the supplemental material.}
    \item Did you specify all the training details (e.g., data splits, hyperparameters, how they were chosen)?
      \answerYes{}
    \item Did you report error bars (e.g., with respect to the random seed after running experiments multiple times)?
      \answerNo{}
    \item Did you include the total amount of compute and the type of resources used (e.g., type of GPUs, internal cluster, or cloud provider)?
      \answerYes{See the supplemental material.}
  \end{enumerate}
  \item If you are using existing assets (e.g., code, data, models) or curating/releasing new assets...
  \begin{enumerate}
    \item If your work uses existing assets, did you cite the creators?
      \answerYes{}
    \item Did you mention the license of the assets?
      \answerYes{}
    \item Did you include any new assets either in the supplemental material or as a URL?
      \answerYes{The code and pretrained models will be released.}
    \item Did you discuss whether and how consent was obtained from people whose data you're using/curating?
      \answerNo{}
    \item Did you discuss whether the data you are using/curating contains personally identifiable information or offensive content?
      \answerNo{}
  \end{enumerate}
  \item If you used crowdsourcing or conducted research with human subjects...
  \begin{enumerate}
    \item Did you include the full text of instructions given to participants and screenshots, if applicable?
      \answerNA{}
    \item Did you describe any potential participant risks, with links to Institutional Review Board (IRB) approvals, if applicable?
      \answerNA{}
    \item Did you include the estimated hourly wage paid to participants and the total amount spent on participant compensation?
      \answerNA{}
  \end{enumerate}
\end{enumerate}

\clearpage

\newpage

\FloatBarrier

\clearpage
\newpage

\appendix
\begin{center}
\Huge\textbf{Appendix}
\end{center}

\section{Limitations}
\label{suppsec:limitations}

While our approach boosts performance of both retrieval-augmented AR and diffusion models and significantly lowers the count of trainable parameters compared to their fully-parametric counterparts, our models still have more trainable parameters than other types of generative models, e.g GANs (Tab.~\ref{tab:uin_metrics}).
Futhermore, we note the long sampling times of both \diffusionmodel and \armodel compared to single step generative approaches like GANs or VAEs.
However, this drawback is inherited from the underlying model class and is not a property of our retrieval-based approach.
Neighbor retrieval is fast and incurs negligible computational overhead.

Another limitation is an inherent tradeoff between database size (and associated storage and retrieval costs) and model performance, as evident from Fig.~\ref{fig:fpp}.
Storing and searching indices for databases of up to billions of images can become quite costly.
Furthermore, our approach depends on the image representation chosen to encode images from the retrieval database $\mathcal{D}$ and the retrieval model.
Both have significant influence on the performance of the RDM/RARM and further research is needed to determine the best choices here.

Our work demonstrates the benefits of adding an external database in general.
However, the choice of the underlying dataset as well as the overall construction strategy of this database is not further investigated.
Sec.~\ref{suppsubsec:patchsize} analyzes the effect of the patch size, yet these patches are chosen randomly and it is an open question for future research if generating patches from the dataset in a systematic way further improves the obtained results.

Finally, this work does not investigate the scaling behavior of semi-parametric generative modeling.
This would be an interesting direction for future work, as we already observe that a model trained only on ImageNet acquires strong zero-shot capabilities, see e.g. Sec.~\ref{subsec:conditional} and \ref{subsec:stylization}, although this dataset is small and obtains limited diversity compared other publicly available datasets~\cite{schuhmann2021laion400m, sharma2018conceptual, DBLP:journals/corr/abs-1811-00982}. Work in NLP ~\cite{retro} suggests that retrieval-augmented transformer models obey a scaling behavior, and we hypothesize that such a property might also exist for image models.
The dependence on the CLIP encoder (e.g. ViT-B/32 vs ViT-L/14) should also be investigated in future work.

\cliptextquerysupp
\suppstylizer
\section{Societal Impacts}
\label{suppsec:societalimpacts}

Large-scale generative image models enable creative applications and autonomous media creation, but can also be viewed as a dual-use technology~\cite{denton2021workshop} with negative implications.
A notorious example are so-called ``deep fakes'' that have been used, for example, to create pornographic ``undressing'' applications~\cite{denton2021workshop}.
Furthermore, the immediate availability of mass-produced high-quality images can be used to spread misinformation and spam, which in turn can be used for targeted manipulation in social media~\cite{denton2021workshop,franks2018sex}.

Datasets are crucial for deep learning as they are the main input of information.
For our model, this concerns the data used in training and inference, as the retrieval database can be considered as a part of the model.
Therefore, the diversity and bias of the synthesized images depends heavily on the diversity and bias in these datasets. For example, a bias of representing a particular skin tone or gender imbalance (i.e., a lack of diversity) already present in the datasets can be easily amplified by deep learning models trained on it~\cite{esser2020note,jain2020imperfect,torralba2011unbiased}; and the effect of post-training truncation models on these phenomena remains under-explored.
However, we note that quantitative diversity analysis of our retrieval-based approach shows that it better covers the data distribution, resulting in less bias towards certain modes in the datasets, such as overrepresented communities, and might be a step towards more balanced and controllable generative models.

Furthermore, one should consider the ability to curate the database to exclude (or explicitly contain) potential harmful source images.
When creating a public API that approach could offer a cheaper way to offer a safe model than retraining a model on a filtered subset of the training data or doing difficult prompt engineering.
Conversely, including only harmful content is an easy way to build a toxic model.

Large-scale image datasets that are used to train advanced synthesis models are usually scraped from the internet \cite{schuhmann2021laion400m, schuhmann2022laion}, and the ethical implications of training on, for example, original digital artwork remain an open question. In addition, it is difficult to assess what impact a single training image had on a generated image or the final generative model.

That is in contrast to the image database used for the retrieval algorithm:
Here, retrieved images have a discernible effect on the output, and the database used during inference may only consist of relatively few high quality images.
Therefore, this could allow for attribution and compensation of the involved content creators.
As an example, when providing an online interface for a retrieval augmented synthesis model, that cost can be factored in together with the hardware costs and be automatically paid for each generated image.
However, the extent to which retrieved representations alone contribute to the final model output needs further investigation.

Lastly, training large image synthesis models with millions of parameters using specialized hardware\footnote{See section:~\ref{suppsubsec:training} for details on the hardware used for the experiments in this work} requires significant financial investment and is therefore available only to a limited number of institutions. The limited access to these large models becomes particularly problematic if these powerful models are not made freely available\footnote{Often a publication of the trained model weights or of the source code is rejected with reference to the dual-use properties listed above.} after training and remain exclusively in the hands of these same institutions, hindering full exploration of their capabilities and biases. %

\section{Concurrent Work}
\label{suppsec:concurrent}
Very recently, two concurrent approaches related to our work, unCLIP~\cite{dalle2} and kNN-Diffusion~\cite{knn-diff}, have been proposed. unCLIP produces high quality text-image results by conditioning a diffusion model on the image representation of CLIP~\cite{clip} and employing large-scale computation. However, unlike our work, it conditions on the CLIP representation of the training image itself, which makes it necessary to learn a generative text-image prior in the CLIP space later.
We show that the neighbor-based approach provides an alternative to training a generative prior to translate between CLIP embeddings (see Sec.~\ref{subsec:conditional} and Fig.~\ref{fig:clipretrosupp}). Our approach allows to modify the retrieval database $\mathcal{D}$ \emph{after} training, which can be used to control the style of the rendered samples (Sec.~\ref{subsec:stylization}, Fig.~\ref{fig:stylizer_supp}). We also show that unCLIP can be interpreted as a special case of our formulation \emph{without} an external database and the retrieval strategy $\sampling(x) = \{\phi_{\text{CLIP}}(x)\}$, for which we train a conditional normalizing flow as the generative prior (see Sec.~\ref{subsec:conditional}).

kNN-Diffusion, like our approach, avoids this problem by conditioning on a neighborhood of the image. Although both kNN-Diffusion and our approach are fundamentally very similar, we use a continuous rather than a discrete diffusion formulation, analyze different forms of neighborhood representations, investigate autoregressive models in addition to diffusion models and are not exclusively limited to text-image synthesis.

\section{Trading Quality for Diversity}
\label{suppsec:qualitydiversity}
Here, we present additional details on top-m sampling and further elaborate on the classifier-free guidance technique for \armodel.

\subsection{Further Details on Top-m Sampling}
\label{suppsubsec:topm_details}
Many approaches to (conditional) generative modeling offer ways to trade off sample quality for diversity at test time. GANs and diffusion models can achieve this by leveraging conditional information via \emph{truncated sampling}~\cite{biggan} and \emph{classifier guidance}~\cite{adm, ho2021classifier},
while models based on a categorical distribution like most autoregressive models allow for \emph{top-k sampling}~\cite{DBLP:journals/corr/abs-1805-04833}.

We propose a similar technique for semi-parametric generative models. Let $Z_m = \sum_{\tilde{x} \in \mathcal{D}^{(m)}} p_{\mathcal{D}}(\tilde{x})$, where $\mathcal{D}^{(m)} \subseteq \mathcal{D}$ is the subset containing the fraction $m \in (0, 1]$ of most likely examples $\tilde{x} \sim p_{\mathcal{D}}(\tilde{x})$.
Similar to top-k sampling, we define a truncated distribution
\begin{equation}
\label{eq:trunc_dist}
\mu(\tilde{x}) =
    \begin{cases}
        p_{\mathcal{D}}(\tilde{x}) / Z_m\, , & \text{if} ~ \tilde{x} \in \mathcal{D}^{(m)} \\
        0 \, , & \text{else} \, ,
    \end{cases}
\end{equation}
which we can use as proposal distribution to obtain $\mathcal{P}$ according to Eq.~\eqref{eq:uncond_sampling}. Thus, for small values of $m$, this yields samples from a narrow, almost unimodal distribution. Increasing $m$ on the other hand, increases diversity, potentially at the cost of reduced sample quality. We analyze this trade-off in Sec.~\ref{subsec:sampling_hacks} and show corresponding visual samples in Fig.~\ref{fig:topm_visual} and Fig.~\ref{fig:topm_visual_rarm}.
In analogy to top-k sampling, we dub this sampling scheme \emph{top-m sampling}.
\topmvisual
\rarmtopmvisual

To gain additional flexibility during inference, this scheme can further be combined with model-specific sampling techniques such as \textbf{classifier-free diffusion guidance}~\cite{ho2021classifier}, since our model \diffusionmodel is a conditional diffusion model of  of the nearest neighbor encodings $\phi(y)$. We present results using different combinations of $m$ and classifier-free guidance scales $s$ in Sec.~\ref{subsec:sampling_hacks}. Moreover, we show accompanying visual examples for the effects of classifier-free unconditional guidance in Fig.~\ref{fig:retro_guiding_samples}.
\retroguidingsamples

\subsection{Classifier-free Guidance for \armodel}
\label{suppsubsec:rarm_guidance}
Classifier-free guidance~\cite{ho2021classifier} was originally proposed for conditional diffusion models, nonetheless, it can also be applied to conditional autoregressive transformers~\cite{ar_cfg}.
We find that, similar to the diffusion head, (\cf Sec.~\ref{subsec:sampling_hacks}) it is sufficient to condition the \armodel on a zero representation to gain an improvement using classifier-free guidance during test time without additional unconditional training.
Given previous image tokens \(t_{1}, \ldots, t_{k-1}\) guidance can then be applied as
\begin{equation}
\begin{split}
    &\log(p_{\text{cfg}}(t_{k} \mid t_{<k}, \{y_{i}\}_{i})) \\
    &\quad= \log(p_{\theta}(t_{k} \mid t_{<k}, \{0\})) + s \cdot \Big( \log(p_{\theta}(t_{k} \mid t_{<k}, \{y_{i}\}_{i})) - \log(p_{\theta}(t_{k} \mid t_{<k}, \{0\}))\Big) \,.
\end{split}
\end{equation}

Qualitative samples obtained with this strategy are depicted in Fig.~\ref{fig:rarm_guiding_samples}.
\rarmguidingsamples

\newpage

\section{Additional Experiments}
\label{suppsec:experiments}

\subsection{Detailed Evaluation on Zero-Shot Stylization}
\label{suppsubsec:ic_comparison}
\subsubsection{Quantitative Evaluation}
\label{suppsubsubsec:ic_quant}
\wikiartmetrics
In this section we quantitatively evaluate the zero-shot stylization capabilities of \diffusionmodels presented in Sec.~\ref{subsec:stylization} and explore their limitations on that task. First, we assess the post-hoc steerability of \diffusionmodels by exchanging the database at inference time and compare it with that of IC-GAN. We use WikiArt~\cite{wikiart} as inference database both for our ImageNet \oidiffusion and for the publicly released IC-GAN trained on ImageNet\footnote{Code and model taken from \url{https://github.com/facebookresearch/ic_gan}} and generate 50K examples with each model. By computing FID, Precision and Recall scores against WikiArt, we can measure how well the two models approximate the WikiArt image manifold. From Tab. \ref{tab:wa_metrics} we can see that \diffusionmodel outperforms IC-GAN on all metrics. Since it better approximates the target image manifold, we conclude that our model can better adopt the properties of this novel database during inference.

\icganclassifier
Furthermore to explicitly compare how well the two models preserve properties of the inference-time database, we train a linear-probe on ResNet-50 features to distinguish between images from these two datasets. The resulting classifier achieves an accuracy of 96\% on an unseen validation set. We measure its accuracy on the 50K generated images for each class from both methods and show the obtained results in Fig.~\ref{fig:ic_classifier}. We see that for both ImageNet and WikiArt, a higher percentage of images generated by \diffusionmodel are classified as belonging to the respective dataset, thus showing \diffusionmodels to better adopt those databases' properties.
\icgangulaschsuppe
\subsubsection{Qualitative Evaluation}
\label{suppsubsubsec:ic_qual}
We also show a qualitative side-by-side comparison between \diffusionmodel and IC-GAN in Fig.~\ref{fig:rdm_vs_icgan} when using the respective train databases (left), the PacsCartoon dataset~\cite{li2017deeper} (mid) and WikiArt (right) as inference database. It shows that \diffusionmodel not only achieve higher visual quality on the task it was trained for, i.e. generation on ImageNet, but also that the generated images based on the novel, exchanged inference contain significantly more properties of the respective databases than those of IC-GAN. However, we also see that \diffusionmodels struggle to generate realistic examples when conditioned \emph{semantic} concepts they have never seen in the training as visible for the 'giraffe' cartoon sample in the fourth column of the bottom row.

\subsection{Alternative Image Encoders $\phi$}
\label{suppsubsec:sp_rep}
As conditioning on raw image pixels would result in excessive memory/storage demands, finding an appropriate compressed representation $\phi(y)$ for the retrieved neighbors
$y \in \mathcal{M}_{\mathcal{D}}^{(k)}$ is of central importance.
For our main experiments we implement $\phi$ with the CLIP image encoder
\nnreps
as its embedding space is compact and shared with the text-embeddings of the CLIP text encoder.
There are principally many other choices of $\phi$ possible, including learning it jointly together with the decoding head. However, since representations pretrained on a large corpus of data has proven not only train-time memory efficient but also beneficial for image generation, we here focus on such pretrained feature extractors.
We investigate two types of representations and compare those from a pre-trained VQGAN encoder~\cite{taming}, representations from the image encoder of CLIP~\cite{clip}. For these experiments we focus on \diffusionmodel, i.e. we implement the decoding heads of the compared models as a conditional diffusion model. For both compared models we use $k = 4$ nearest neighbors during training and inference. Moreover we compare them with a full-parametric \emph{LDM} baseline with 1.3$\times$ more parameters. To render training less compute intensive, %
we train them on the ImageNet-dogs subset (see Sec.~\ref{suppsubsubsec:in_subset_stats}).

Fig.~\ref{fig:nn_rep} summarizes the obtained results which again demonstrate the efficacy of semi-parametric generative modeling compared to fully-parametric models, as both VQGAN-\footnote{For more details on how we feed this representation to the decoding head via cross attention, see Sec.~\ref{suppsubsubsec:vq_sequence}.} and CLIP-nearest-neighbor encodings improve sample quality (higher precision~\cite{DBLP:journals/corr/abs-1904-06991}, lower FID~\cite{FID}) as well as diversity (higher recall~\cite{DBLP:journals/corr/abs-1904-06991}), despite using less trainable parameters (the baseline uses $1.3 \times$ more parameters). Moreover we see that the model conditioned on CLIP image embeddings consistently improves over that which uses VQGAN encodings. Thus we use such models for our experiments in the main paper.

\subsection{Patch Size of Images in the Database}
\label{suppsubsec:patchsize}

Our retrieval database consists of 20M examples originating from the OpenImages~\cite{DBLP:journals/corr/abs-1811-00982}, see Sec.~\ref{suppsubsec:retrieval} for details. As the images in OpenImages are much larger than our train time image size of 256 pixels per side, we crop multiple patches per image. For the train database used for the models presented in the main experiments we use a patch size of $256 \times 256$ pixels. However, since the chosen patch size determines the properties of the images in the database\footnote{Larger patch sizes will result in more images depicting objects as a whole, whereas smaller patch sizes will rather show object parts.} we investigate the effects of varying the size of the extracted patches in the database.
\patchsizefig
To this end we train three identical \diffusionmodel with $k=4$ with databases consisting of patches which were extracted from OpenImages by using different patch size $H_{\mathcal{D}} = W_{\mathcal{D}} \in \{64, 128, 256\}$. As in Sec.~\ref{suppsubsec:sp_rep} we train the models on the dogs-subset of ImageNet compare the semi-parametric models with an \emph{LDM} baseline with 1.3$\times$ more trainable parameters.

Fig.~\ref{fig:patchsize} visualizes the obtained results.
Vertical and horizontal bars denote the performance of the \emph{LDM} baseline.
As expected, we observe the patch size to have substantial influence on the performance of semi-parametric models.
We see that an patch size of 64 pixel seems to be too small, resulting in worse performance compared to the baseline.
Increasing the patch size results in significant improvements over the baseline, despite a smaller parameter count.
High precision~\cite{DBLP:journals/corr/abs-1904-06991} and FID~\cite{FID} indicate that conditioning on larger patches results in improved sample quality. Recall values~\cite{DBLP:journals/corr/abs-1904-06991} decrease when increasing the patch size. This is due to the fact that for the model with a patch size of 64, the generated samples lack perceptual consistency, as indicated by the small precision values. However the model with a database of patch size 256 still has a recall score $> 0.60$ which is still high and clearly larger than the achieved value of the baseline.
This demonstrates that retrieval-augmented models maintain high sample diversity and conditioning on global object attributes yields more coherent samples than only using local object parts in the database. In the future, increasing the patch size beyond 256 px per side bears potential to further improve sample quality achieved by semi-parametric models.%

\subsection{Optimizing $k_{\text{\tiny train}}$ for \armodel}
\label{suppsubsec:knn_rarm}
\rarmknnwrapped
Similar to Sec.~\ref{subsec:unconditional} we here evaluate suitable choices of $k_{\text{\tiny train}}$ for \armodel and therefore train models with the same decoding head but with different $k_{\text{\tiny train}} \in \{1,2,4,8,16\}$ on the ImageNet dogs subset. We show the resulting evaluation metrics computed based on 2000 samples in Fig.~\ref{fig:rarm_knn_wrapped}, where we observe the models with $k_{\text{\tiny train}} \in \{ 2, 4\}$ to perform best as both models yield good FID scores while sill achieving comparably high precision and recall values. The optimal choice seems to be $k_{\text{\tiny train}} = 2$ which is different than for \diffusionmodel, where we found $k_{\text{\tiny train}} = 8$ to yield the best results. %

\subsection{Top-m Sampling for \armodel}
\label{suppsubsec:rarm_topm}
\rarmtopmtruncationwrapped
In this section we analyze the effects of top-m sampling for \armodel similar to the evaluation for \diffusionmodel presented in Fig.~\ref{fig:topm_trunc_wrapped}. To this end we use the best performing model for $k_{\text{\tiny train}} = 2$ from Sec.~\ref{suppsubsec:knn_rarm} and generate 10000 samples for $m \in \{1., 0.5, 0.1, 0.01, 0.001, 0.0001\}$ without classifier-free guidance. Fig.~\ref{fig:topm_trunc_rarm} visualizes the results which show the same truncation behavior as observed for \diffusionmodel, see Fig.~\ref{fig:topm_trunc_wrapped}, including the FID-IS sweet spot at $m=0.01$. This experiment provides further evidence for the discussed advantages of semi-parametric generative models compared to their fully-parametric counterparts, irrespective of the actual realization of the decoding head.

\section{Implementation Details}
\label{suppsec:implementation_details}

\subsection{Train-time Database and Retrieval Strategy}
\label{suppsubsec:retrieval}
As mentioned in the main paper, we build our database from the OpenImages dataset~\cite{DBLP:journals/corr/abs-1811-00982}  , which contains 9M images of varying spatial sizes with a shorter edge length of at least 1200 px. To build our 20 M images database we resize all images such that the shorter edge length is equal to 1200 px and subsequently randomly select 2-3 patches of size $256 \times 256$ px per image of OpenImages. Thus, we use each of these images at least once. We investigate the effects of using different patch sizes for building the database in Sec.~\ref{suppsubsec:patchsize}.

For all datasets investigated in the work, we precompute $k=20$ neighors for each query image of a given train dataset and store the resulting CLIP-embeddings along with the image ids and patch coordinates of the corresponding image in the OpenImages dataset. This allows us to also visualize the images corresponding to the neighbors in the CLIP space.

For nearest neighbor retrieval we use the ScaNN search library~\cite{pmlr-v119-guo20h}. With this choice, retrieving 20 nearest neighbors from the database described above takes approximately 0.95 ms. Thus, including NN retrieval in the training process would also not mean significant training time overheads.

\subsection{Training Details}
\label{suppsubsec:training}

\subsubsection{Models with Diffusion Based Decoding Heads}
\label{suppsec:implementation_details_rdm}
In Tab.~\ref{tab:hyperparams} we show the hyperparameters which were used to train our presented models, which use diffusion based decoding heads. For the retrieval-augmented models, the hyperparameters correspond only to the decoding head, as the other parts of the model are not trainable. We trained our main model (which was used to generate all qualitative results in this work as well as the quantitative results shown in Tab.~\ref{tab:uin_metrics}) on eight NVIDIA A-100-SXM4 with 80GB RAM per GPU. The overall training time compute spent to train this model is 48 A-100 days when considering a single A-100 with 80 GB RAM or 96 A-100 days when calculating with an A-100 with 40GB.

The models evaluated in the $k_{\text{\tiny train}}$ experiments in Sec.~\ref{subsec:unconditional} and Sec.~\ref{subsec:conditional} are all trained on two NVIDIA A-100-SXM4 with 80GB RAM per GPU for the same number of train steps. To enable larger batch size we only parameterize these models with 200M trainable parameters and use a compression model which is trained with KL-redularization with a downsampling factor $f = 16$. For a detailed explanation of the compression models and of the \emph{LDM} framework, see~\cite{ldm}. This is in contrast to our other diffusion based models, which use a VQ-regularized compression model with $f=4$, as $f=16$ allows us to further increase the batch size and thus result in faster converging models.
The normalizing flow used to model the CLIP generative prior in Sec.~\ref{subsec:conditional} uses a ``modernized'' version of the invertible backbone, built from 200 blocks that consist of coupling layers~\cite{dinh2016density}, activation normalization~\cite{glow} and shuffling as in ~\cite{DBLP:conf/nips/RombachEO20,Blattmann_2021_ICCV}. We replace batch normalization in the sub-networks with layer normalization and RELU with GELU~\cite{gelu} nonlinearities.

The models from the analysis using the subsets of ImageNet in Sec.~\ref{subsec:dataset_complexity} are all trained on a single NVIDIA A-100 GPU with 40GB RAM. To be able to use the same batch sizes also for the \emph{LDM} baselines shown in these experiments, each of which has 1.3 times more parameters than the corresponding \diffusionmodel, we use gradient checkpointing~\cite{Chen2016TrainingDN} to reduce memory cost during backpropagation at the expense of additional computations in the forward pass. As these baselines are 'common' unconditional models, we use self-attention (SA) instead of the cross-attention layers (CA) which are used to feed the nearest neighbor representation $\phi$ to the decoding head of the semi-parametric models. All our models are implemented in PyTorch. We will release the code and pretrained models in the near future.

\hyperparams

\subsubsection{Models with Autoregression Based Decoding Heads}
In Tab.~\ref{tab:hyperparamsRARM} we show the hyperparameters which were used to train the autoregressive models presented in this work.
For the retrieval-augmented models, the hyperparameters correspond only to the decoding head, as the other parts of the model are not trainable.
All autoregressive models are decoder-only GPT-like transformer models and use the same VQGAN compression model with a downsampling factor of \(f=16\). Using such a compression model and applying raster scan reordering~\cite{vqvae} results in an input sequence of length $256$ for an image of spatial size $256 \times 256$. This prevents our models from allocating excessive amounts of GPU memory, what can arise for long sequences, due to the quadratic complexity of the attention mechanism.
The \armodel have an additional cross-attention block (CA) behind every self-attention block that is used to feed the nearest neighbor representation \(\phi\) to the decoding head.
We train all autoregressive models on a single NVIDIA A-100 with 40GB RAM.

\hyperparamsRARM

\vspace{1cm}
\subsubsection{Statistics for ImageNet subsets}
\label{suppsubsubsec:in_subset_stats}
\imagenetsubsets
In Tab.~\ref{tab:in_subsets} we present detailed statistics for the datasets involved in the comparison of fully- and semi-parametric generative models for increasing complexity of the modeled data distribution. For the dogs subset, we used the class labels ranging from 181 to 280, resulting in a training dataset containing $N =$ 163K examples. Including all mammals lead to overall 241 classes with $N =$ 309K examples whereas training on all 398 classes referring to animals resulted in a dataset of $N =$ 511K individual images. As for our main experiments, we did not use any class labels for training the models on these datasets.

\subsection{Evaluation Details}
\label{suppsubsec:eval}
\subsubsection{Analysis Experiments on Effects of $k_{\text{\tiny train}}$ from Sec.~\ref{subsec:unconditional}}
To generate the results shown in the $k_{\text{\tiny train}}$ analysis presented Sec.~\ref{subsec:unconditional} we used $m=0.01$ and no guidance for all compared choices of $k_{\text{\tiny train}}$. we assessed performance metrics based on 1000 samples for each run.

\subsubsection{Comparison with State of the Art}
For the SOTA comparison presented in Sec.~\ref{subsec:unconditional}, we use the evaluation protocol proposed in ADM~\cite{adm}, where performance metrics are calculated based on 50K samples and by using the ImageNet train set as a reference for the data distribution. We also use their publicly available evaluation implementation to obtain comparable results\footnote{\url{https://github.com/openai/guided-diffusion}}. To be able to compare our models also with IC-GAN~\cite{ic-gan}, which uses train set instances during evaluation, we additionally follow their protocol of evaluating against the validation split. Moreover, we compute precision and recall scores for their method, by using the publicly available pretrained weights\footnote{\url{https://github.com/facebookresearch/ic_gan}} for both train and validation splits, see Tab.~\ref{tab:uin_metrics}. The low recall scores indicate their generated samples to lack diversity and their GAN based model to only capture few modes of the data distribution, which is a well-known issue for GANs~\cite{srivastava2017veegan,wassersteingan, unrolledgan,pacgan}. In contrast, since our models profit from the mode-covering property of the likelihood based objective, our recall scores are sufficiently high for all presented combinations of sampling parameters.

\subsubsection{Details on Evaluations on Text-to-Image Generalization}
\label{suppsubsubsec:tesx2img_generalization}
In Sec.~\ref{subsec:conditional} we evaluate the the generalization capabilities of our ImageNet \diffusionmodel, which is trained only on images, when applied to text-to-image synthesis. For generating the ImageNet-FIDs presented in Fig.~\ref{fig:generalization_quant} we used 2000 samples generated with $\text{top-m} = 0.01$ and without unconditional classifier-free guidance. The presented scores for text-to-image synthesis on COCO were synthesized with $\text{top-m} = 0.01$ and classifier-free guidance scale $s = 2.0$ for all models. We furthermore applied the same sampling parameters when generating results with the model directly conditioned on CLIP representation, which includes a flow prior for closing the mismatch between CLIP text- and image-embeddings. %

\subsubsection{Details on Experiments regarding Dataset Complexity}
\label{suppsubsubsec:dset_complexity}
For both \diffusionmodel and \armodel we compute the metrics presented in Fig.~\ref{fig:dataset_complexity} based on 1000 samples for each individual dataset and use $k = 4$. We also compute the metrics for the fully-parametric baselines with 1000 samples. For \diffusionmodel, we use $\text{top-m} = 0.01$ and no classifier-free guidance. For \armodel we use \(\text{top-m} = 0.005\), top-\(k = 2048\) and no classifier-free guidance.

\subsubsection{Building a Conditioning Sequence with VQGAN-encodings}
\label{suppsubsubsec:vq_sequence}
In the comparison regarding different encoders $\phi$ in Sec.~\ref{suppsubsec:sp_rep} we compare CLIP image embeddings with those extracted by a pretrained VQGAN encoder. However, for the latter, which yields a three-dimesional tensor for each retrieved nearest neighbor, we have to apply a reshaping to obtain a sequence, which is suitable for being fed to the decoding head via cross-attention. We here implement $\phi$ with a $f16$ VQGAN-encoder pretrained on OpenImages~\cite{ldm}\footnote{\url{https://github.com/CompVis/latent-diffusion}}. For the default VQGAN input size, which is 256, the latent code of each retrieved neighbor would be of size $16 \times 16 \times 256$. Thus, to further shrink to dimensionality of this representation we resize the input images for each of the $k = 4$ nearest neighbors to $128 \times 128$ px, since this does not hurt the model's performance, resulting in a latent tensor of shape $ 8 \times 8 \times 256$. We then form a sequence shape $64 \times 256$ for each nearest neighbor representation by applying raster scan reordering~\cite{vqvae} and subsequently concatenate all $k=4$ individual representation channel-wise, resulting in the final conditioning input for the decoding head with a shape of $64 \times 1024$ which can be fed via cross attention.

\section{Additional Samples}
\label{suppsec:additional_samples}
In this section we show additional qualitative samples for all presented experiments in
the main paper. Fig.~\ref{fig:clipretrosupp} shows additional samples of the generalization of our ImageNet \diffusionmodel, when using CLIP-representations of text prompts as inputs, as in Fig~\ref{fig:clipretro}. Fig.~\ref{fig:stylizer_supp} shows additional examples of text-guided stylization with by changing the database for the model ImageNet model mentioned above. With this zero-shot stylization model, we can also generate unconditional samples. This is visualized in Fig.~\ref{fig:random_styles_comp} and compared with unconditional samples from the same model, with the original database $\mathcal{D}_{\text{\tiny train}}$, which is used during training. We furthermore show additional unconditional samples in Fig.~\ref{fig:uin_rsamples} and also more class-conditional samples similar to Fig.~\ref{fig:zero_shot_cc} in Fig.~\ref{fig:more_cc_samples}. Additional samples from our experiment which compares the direct use of CLIP text-embeddings and embeddings from a conditional normalizing flow (as in Sec.~\ref{subsec:conditional}) are depicted in Fig.~\ref{fig:generatorsuppe}.
Random samples from the autoregressive models are shown in Fig.~\ref{fig:rarm_rsamples}.

\databasecomp

\uinrandomsamplestwo

\rarmrandomsamplestwo

\moreccsamples
\generatorsuppe

\ffhqrandoms

\end{document}